# Assessing Wildfire Susceptibility in Iran: Leveraging Machine Learning for Geospatial Analysis of Climatic and Anthropogenic Factors

Ehsan Masoudian[1], Ali Mirzaei[2], Hossein Bagheri[2*]

## Abstract

This study investigates the multifaceted factors influencing wildfire risk in Iran, focusing on the interplay between climatic conditions and human activities. Utilizing advanced remote sensing, geospatial information system (GIS) processing techniques such as cloud computing, and machine learning algorithms, this research analyzed the impact of climatic parameters, topographic features, and human-related factors on wildfire susceptibility assessment and prediction in Iran. Multiple scenarios were developed for this purpose based on the data sampling strategy. The findings revealed that climatic elements such as soil moisture, temperature, and humidity significantly contribute to wildfire susceptibility, while human activities—particularly population density and proximity to powerlines—also played a crucial role. Furthermore, the seasonal impact of each parameter was separately assessed during warm and cold seasons. The results indicated that human-related factors, rather than climatic variables, had a more prominent influence during the seasonal analyses. This research provided new insights into wildfire dynamics in Iran by generating high-resolution wildfire susceptibility maps using advanced machine learning classifiers. The generated maps identified high-risk areas, particularly in the central Zagros region, the northeastern Hyrcanian Forest, and the northern Arasbaran forest, highlighting the urgent need for effective fire management strategies.

---

[*] Faculty of Civil Engineering and Transportation, University of Isfahan, Isfahan, Iran. Correspondence: h.bagheri@cet.ui.ac.ir

[1] School of Surveying and Geospatial Engineering, College of Engineering, University of Tehran, Tehran, Iran. ehsan.masoudian@ut.ac.ir (Ehsan Masoudian)

[2] Faculty of Civil Engineering and Transportation, University of Isfahan, Isfahan, Iran. alimirzaei1109@trn.ui.ac.ir (Ali Mirzaei)

# 1  Introduction

Forests are essential for maintaining biodiversity, regulating the climate, mitigating droughts and floods, absorbing atmospheric carbon dioxide, and providing crucial food and medicinal resources (Morales-Hidalgo et al., 2015; Qiu et al., 2020; White et al., 2016). However, wildfires pose significant threats to these vital ecosystems. Advanced technologies like remote sensing and spatial data analysis offer effective methods for monitoring and identifying areas at high risk of wildfires. With its continuous spatial and temporal coverage, remote sensing is a cost-effective and precise tool for assessing the spatial-temporal dynamics of wildfire outbreaks. This capability enables researchers and environmental authorities to understand better and manage the factors influencing wildfire occurrences and impacts (Adab et al., 2013; Cetin et al., 2023; Chakraborty et al., 2024; Eskandari & Chuvieco, 2015).

A key aspect of identifying wildfire-prone areas involves analyzing the various factors that contribute to fire ignition and spread (Jahdi et al., 2020; Jaiswal et al., 2002). These include climatic conditions such as temperature and humidity, vegetation traits like tree density, human activities such as cooking with fire or producing charcoal, urban development near forests, proximity to roads or power lines, and topographical features.

Previous studies have extensively examined the influence of various parameters on forest fires. For example, Pramanick et al. (2023) analyzed forest fires in a region of India using a linear regression model to evaluate the relationship between topographic information, road networks, and remote sensing data with wildfire occurrences. While their study provided valuable insights, it is noteworthy that meteorological data, which play a significant role in wildfire occurrence, were not incorporated into their proposed model (Pramanick et al., 2023).

A recent study utilized topographical, meteorological, remote sensing, and spatial data to identify fire-prone areas in the western Himalayan forests by applying a Fuzzy method. While the study accounted for several critical factors, the evaluations demonstrated the limited generalizability of the developed model to other regions (Pragya et al., 2023).

Shao et al. (2022) developed a forest fire risk map for China using topological information, climate data, and machine learning models, offering a quantitative analysis of the impact and importance of each parameter with promising results. However, the study did not consider human factors (Shao et al., 2022).

Similarly, another study investigated the factors influencing fire, including topological features and changes in forest cover over time, within a specific region of North Korea (Jin & Lee, 2022).

Saidi et al. (2021) conducted a case study on a forest fire in a region of Tunisia, highlighting a strong correlation between spatial information layers, remote sensing data, and forest fire occurrences. However, the study did not provide an individual analysis of the impact of each parameter (Saidi et al., 2021).

Yang et al. (2023) used Landsat 7 data and a CNN model to identify fire-prone areas in Indonesian forests, aiming to determine the non-linear relationship between various Landsat 7 image bands and fire occurrence through algorithm development. However, the study did not consider the role of topographic parameters, meteorological factors, and spatial information layers (Yang et al., 2023).

In another study, a forest fire risk map was developed for two regions in Nepal, utilizing land use metrics, temperature, topographic information, and distance from main roads as parameters influencing forest fires. The researchers employed a linear regression model to identify fire-prone areas. However, the model's ability to capture the non-linear relationships among specific parameters affecting fire risk estimation was limited (Parajuli et al., 2020).

A study conducted in the Gachsaran region of Iran used remote sensing data, spatial information, and topographic and meteorological data to identify high-risk fire points. Multi-criteria decision-making (MCDM) methods were employed to assign weights to these parameters. However, the study's results were based on a small area, requiring further validation before they can be generalized to other regions (Feizizadeh et al., 2023).

Eslami et al. (2021) investigated fire risk in the forests of northern Iran by integrating remote sensing data and spatial information layers. However, the study did not evaluate the importance of individual input parameters in identifying fire-prone areas. Moreover, the research focused on a limited section of Iran's forests with similar geographical and climatic conditions, raising concerns about the generalizability and applicability of the developed model to other regions (Eslami et al., 2021).

Previous studies in various regions have overlooked critical aspects of identifying wildfire-prone areas. Notably, some studies have not evaluated the performance of advanced models for creating wildfire risk maps, relying instead on simplistic methods such as linear regression, as seen in research conducted in Northern Iran and Nepal (Parajuli et al., 2020; Pragya et al., 2023; Pramanick et al., 2023; Saidi et al., 2021). Despite their low computational demands, simple methods struggle to accurately model the nonlinear and complex relationships among various parameters, resulting in lower accuracy and reliability of the maps. Additionally, some studies have not investigated the importance and influence of input parameters on wildfire risk, which can lead to an incomplete understanding of the factors affecting this risk. These studies have not fully utilized the value of spatial, satellite, topographic, and meteorological data, limiting the comprehensive understanding of wildfire risk factors. Furthermore, many studies have limited their scope to areas with uniform forest textures, often neglecting diverse geographical and climatic conditions, compromising their findings' generalizability.

Iran, as a country with diverse climates, has many areas at risk of wildfires due to a combination of natural factors such as climate, geography, and varied topographies. These wildfires threaten forests, grasslands, shrublands, and urban interfaces (Eskandari et al., 2021) Figure 1 : The frequency of wildfire incidents in Iran in different years, according to the Iranian Statistics Center reports. illustrates the frequency of wildfires in Iran as reported by the Iranian Statistics Center (*Iranian Statistics Center*, 2024). It vividly depicts a substantial number of fire incidents occurring annually, with a marked and alarming escalation observed in the last five years.

Previous studies on wildfires in Iran have predominantly concentrated on specific forest regions, often yielding fragmented findings that hinder the generalization of results on a national scale. Additionally, there has been a notable absence of comprehensive evaluations addressing the diverse factors contributing to wildfire occurrences. To bridge these gaps, this paper investigates the multifaceted drivers of wildfire risk in Iran, focusing on human activities and climatic influences. Leveraging data from advanced platforms like Google Earth Engine and Overpass Turbo, the study provides valuable insights into vegetation patterns, weather dynamics, and environmental fluctuations shaping wildfire susceptibility across the country.

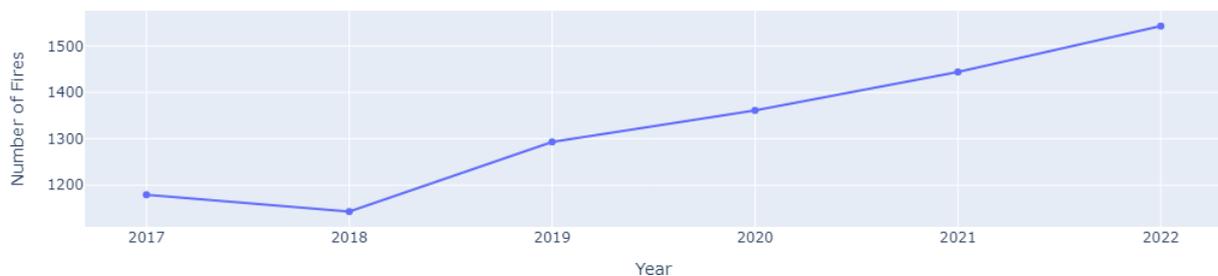

Figure 1 : The frequency of wildfire incidents in Iran in different years, according to the Iranian Statistics Center reports.

Additionally, several machine learning models have been developed and trained on diverse datasets tailored to different scenarios, ensuring accuracy and adaptability. By intentionally masking certain regions, the generalizability of the findings can be assessed, enabling predictions in previously unobserved areas. Moreover, this study incorporates a seasonal analysis to examine how various parameters influence wildfire risk across different times of the year. Recognizing that wildfire risk is dynamic and shaped by seasonal variations, the study evaluates how factors such as temperature, precipitation, and human activities impact wildfire risk during the warm and cold seasons. Finally, the significance of different input parameters has been evaluated, resulting in a comprehensive wildfire risk map for Iran. This map integrates spatial, remote sensing, and meteorological data, highlighting areas requiring immediate attention.

This paper consists of several sections. In the current section, we review previous investigations and outline the paper's main objectives. Section 2 presents the study area's specifications. In Section 3, we describe the methods for wildfire risk mapping, the datasets used for this purpose, and the techniques applied to identify the role of climatic and anthropogenic parameters. The results from various experiments and an evaluation of the proposed method's effectiveness in mapping wildfire risk over Iran are presented and discussed in Sections 4 and 5, respectively.

## 2    Study area

The study area of this research is Iran, located in the southwest of Asia. As illustrated in Figure 2, Iran encompasses approximately 1.6 million square kilometers and is home to over 80 million people (Talebi et al., 2014). Iran's vegetation is notably diverse, influenced by varying climatic and geographical conditions. It ranges from temperate forests to desert ecosystems. Wildfires occur across various vegetation types but are predominantly associated with forested regions.

The country's forest cover, which boasts high biodiversity, can be categorized into four main types: Hyrcanian forests, Arasbaran forests, Zagros forests, and Hara forests (Talebi et al., 2014). Hyrcanian forests, primarily found in the northern provinces of Iran, including Gilan, Mazandaran, and Golestan, consist mainly of oak, fir, and beech trees. These forests thrive in northern Iran's moist, cool climatic conditions, fostering diverse flora and fauna. However, human activities significantly con- tribute to wildfires in this region, exacerbated by complex topography and dense vegetation, which facilitate rapid fire spread (Jaafari et al., 2017). The Arasbaran forests, found in northwestern Iran (West and East Azerbaijan), are home to many rare plants that need particular protection. Their unique biodiversity demands focused conservation efforts.

The Zagros forests, spread across western, southwestern, and southern Iran, are home to oak trees that have adapted to warm, arid conditions. Unfortunately, the rise in intentional fires for making charcoal poses a severe threat to these forests, highlighting the need for action. Lastly, the Hara forests are situated along Iran's southern margins, adjacent to the Persian Gulf and the Sea of Oman.

Forest destruction due to wildfires is a severe environmental issue in Iran (Novo et al., 2024). Wildfires can happen naturally or be caused by human activities, leading to significant damage to forests and their ecosystems (Jahdi et al., 2020). The main factors contributing to forest wildfires in Iran include:

- Climatic Conditions: Climate plays a crucial role in wildfires' intensity and widespread nature. During long periods of drought, forests in Iran become more prone to catching fire, and wildfires can spread quickly. Higher temperatures and lower humidity also raise the risk of wildfires. In contrast, spring rains (especially in western Iran) encourage the growth of underbrush. These plants dry out during the hot season due to lack of rainfall, providing plenty of fuel for wildfires to proliferate.

- Human Activities: It is noteworthy that many wildfires are preventable and frequently result from human activities. The improper use of fire for cooking, excessive charcoal production from forest wood, and poor monitoring of areas near villages and roads are all human behaviors that can lead to devastating wildfires.

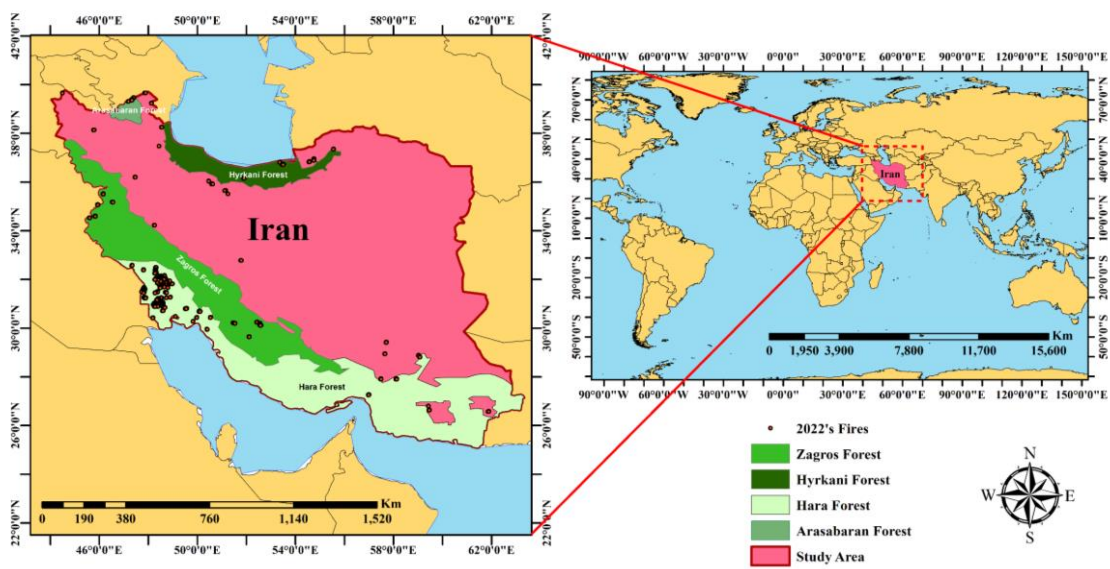

Figure 2: A display of the study area, including the locations of wildfires in 2022 (black circles) and the extent of forests in Iran, highlighted in green.

# 3    Materials and Methods

## 3.1    Study Datasets

This research utilized various remote sensing data, meteorological parameters from climate models such as the Famine Early Warning Systems Network Land Data Assimilation System (FLDAS), and spatial information derived from GIS layers to identify wildfire-prone areas in forests. Some parameters, such as high temperatures and low humidity, may act as fire initiators. In contrast, others, including topographic slope, aspect, dry vegetation, and some human activities, can facilitate fire spread (Adab et al., 2013). These parameters are systematically considered together to enhance the accuracy of identifying wildfire-prone areas. This study aimed to develop high-resolution maps to identify wildfire-prone areas, prioritizing data sources with high spatial resolution. Due to the study's broad spatial and temporal scope, covering all of Iran from 2015 to 2022, datasets were carefully selected for their spatial resolution, temporal compatibility, and full coverage. As no single dataset met all requirements, multiple datasets were integrated to account for Iran's diverse climatic zones. This approach ensured comprehensive data coverage despite the challenges posed by the study's sizeable geographical scope.

Due to the large study area and examining various influential fire factors—climatic conditions, regional topography, and human activities—cloud processing platforms including Google Earth Engine and Overpass Turbo were employed to extract relevant parameters from multiple sources.

Google Earth Engine (GEE) is a cloud-based platform that provides access to a vast geospatial data repository, including satellite imagery and environmental datasets. It offers powerful tools for processing and analyzing large-scale geospatial information, making it invaluable for studies on wildfire susceptibility. GEE's extensive archive of remote sensing data, such as Landsat and MODIS satellite products, allows researchers to monitor vegetation health, land cover changes, and climate variables over time, providing key insights into the conditions contributing to wildfire risk.

Similarly, Overpass Turbo complements GEE by offering access to detailed geospatial data from OpenStreetMap (OSM). This web-based tool, which uses the Overpass QL query language, enables researchers to extract specific data layers, such as vegetation cover, infrastructure, and environmental features, essential in wildfire susceptibility analysis (Barrington-Leigh & Millard-Ball, 2017). By integrating Overpass Turbo's anthropogenic and infrastructural data—such as information on roads, residential areas, and land-use patterns—with climatic and vegetation data from GEE, researchers can achieve a more comprehensive understanding of wildfire risks. This combined approach provides a robust geospatial framework for identifying how human activities intersect with natural factors, influencing fuel availability and ignition sources. Together, these tools enable targeted mitigation strategies and effective management of wildfire-prone areas.

This study utilized the GEE platform to query and retrieve geospatial data related to climatic, topographic, and vegetation parameters. The temporal aggregation process involved calculating each parameter's average or median values from 2015 to 2022 to ensure consistency across datasets. The mean was used for parameters with a concentrated distribution (e.g., precipitation, wind speed), while the median was applied to those with a more dispersed distribution (e.g., soil moisture, humidity). This approach harmonized the datasets and enabled their integration, accounting for variations in temporal coverage and data quality.

Overpass Turbo was also employed to query and extract human-related factors, such as power line proximity and population density. The retrieved data were exported and Euclidean distance and density calculations were performed to investigate the spatial relationships between these factors and wildfire risks. This platform enabled the generation of precise geospatial layers representing human activities and their potential influence on wildfire susceptibility within the study area. Table 1 presents the parameters used to identify wildfire-prone areas, detailing their sources, cloud platforms, and temporal and spatial resolutions. Further discussions on each parameter are provided in the following.

### 3.1.1    Topographic parameters

- **DEM:** The Digital Elevation Model (DEM) provides detailed data on land surface elevation, which is crucial for analyzing and predicting wildfires. The elevation model significantly influences the path and behavior of fires. Higher altitudes may increase the risk of wildfires due to stronger winds (Ahmad et al., 2018; Kanga & Singh, 2017; Novo et al., 2024). In this study, we utilized the DEM derived from the Shuttle Radar Topography Mission (SRTM) sensor, which offers a spatial resolution of 30 m. This data enhances our understanding of terrain characteristics influencing wildfire occurrences in the study area.

- **Slope:** Slope information is critical for assessing wildfire-prone areas, as it influences the speed and direction of fire movement. Steep slopes can accelerate the upward spread of fire, potentially covering extensive areas rapidly. Additionally, controlling fires on steep slopes presents significant challenges due to faster fire spread and limited accessibility (Ahmad et al., 2018; Kanga & Singh, 2017; Novo et al., 2024). This study used slope data derived from SRTM DEM, with a spatial resolution of 30 m, to analyze terrain characteristics affecting fire behavior. This dataset aids in identifying and evaluating areas susceptible to wildfires.

Table 1: Influential parameters employed for wildfire susceptibility and risk assessment in Iran

| Type of Feature | Feature (Notation) | Source | Spatial Resolution | Temporal Resolution |
|---|---|---|---|---|
| **Topographic Parameters** | DEM (SRTM DEM) | SRTM | 30 m | Static |
| | Slope (SRTM Slope) | SRTM | 30 m | Static |
| | Aspect (SRTM Aspect) | SRTM | 30 m | Static |
| **Vegetation Parameters** | EVI (EVI VIIRS) | VIIRS | 500 m | 16 Day |
| | NDVI (NDVI VIIRS) | VIIRS | 500 m | 16 Day |
| **Climatic Parameters** | Precipitation (ERA5 Precipitation) | ERA5 | 11132 m | Daily |
| | Temperature (FLDAS Temp) | FLDAS | 11132 m | Monthly |
| | Humidity (FLDAS Humidity) | FLDAS | 11132 m | Monthly |
| | Wind Speed (FLDAS Wind) | FLDAS | 11132 m | Monthly |
| | Evapotranspiration | MODIS | 500 m | 8 Day |
| | Land Surface Temperature (MODIS LST) | MODIS | 1000 m | Daily |
| | Land Cover Type (MODIS Land Cover) | MODIS | 500 m | Yearly |
| | Soil Moisture (SMAP Soil Moisture) | SMAP | 10000 m | 3 Day |
| **Anthropogenic Parameters** | Major Road Euclidean Distance (Major Road Euc) | Overpass Turbo | - | static |
| | All Road Euclidean Distance (All Road Euc) | Overpass Turbo | - | static |
| | Population Center Euclidean Distance (Population Euc) | Overpass Turbo | - | static |
| | Power Line Euclidean Distance (Power Line Euc) | Overpass Turbo | - | static |
| | Population Density | Overpass Turbo | - | static |
| **Target Parameters** | Fire Mask | MODIS Terra Thermal | 1000 m | Daily |

- **Aspect:** The terrain aspect is crucial for identifying fire-prone areas, as it primarily determines sunlight exposure on the surface. Variations in aspect lead to differing temperatures and humidity levels. Slopes facing the sun before noon tend to be drier and more susceptible to wildfire. In contrast, slopes with orientations that receive less direct sunlight generally maintain higher humidity levels, thereby reducing fire risk (Ahmad et al., 2018; Kanga & Singh, 2017; Novo et al., 2024). In this study, aspect data was derived from SRTM DEM, which provided information at a spatial resolution of 30 m.

### 3.1.2 Vegetation parameters

- **EVI:** The Enhanced Vegetation Index (EVI) is a valuable tool for identifying wildfire-prone areas. Healthy and thriving vegetation typically correlates with positive and high EVI values. Analyzing changes in EVI across forested regions helps identify vegetation dynamics. Areas with a sharp decrease in EVI may indicate past fire occurrences, suggesting an increased risk of future fires in nearby locations (Doan et al., 2024; Sunar & Özkan, 2001; Vasilakos et al., 2018). The VIIRS sensor, which has a spatial resolution of 500 m, provides data for this index.

- **NDVI:** The Normalized Difference Vegetation Index (NDVI) is computed by comparing the reflectance values of the red and near-infrared bands. It yields lower values in areas with dry or dead plant matter than healthy vegetation. Consequently, areas with lower NDVI values are typically more susceptible to fire than those with higher values. This study utilized NDVI data derived from VIIRS sensor imagery, which had a spatial resolution of 500 m (Doan et al., 2024; Sunar & Özkan, 2001; Vasilakos et al., 2018).

### 3.1.3 Climatic parameters

- **Precipitation:** Precipitation significantly affects the probability of wildfires. Low rainfall leads to drier vegetation, increasing the availability of fuel and raising the probability of fire occurrence and spread. Conversely, abundant rainfall maintains soil and vegetation moisture, reducing fire risks. However, lightning, which often accompanies areas of frequent rainfall, poses a significant fire risk in forests. The impact of precipitation on forest fires should be examined in relation to other factors. Monitoring precipitation anomalies in various regions is crucial, as areas with higher precipitation anomalies are more prone to lightning and subsequent wildfire incidents (Parvar et al., 2024; Stanimir et al., 2020). This study utilized daily precipitation data obtained from the ERA5 model, which had a spatial resolution of 1132 m.

- **Temperature:** Temperature is crucial in creating favorable conditions for fire spread in wildfire-prone areas. Higher temperatures contribute to environmental drying, impacting vegetation, including grasses and other underbrush, between trees. This dryness increases the flammability of the surroundings, making combustible materials more likely to ignite. This study utilized daily average temperature data obtained from the FLDAS model (Parvar et al., 2024; Stanimir et al., 2020).

- **Humidity:** Air humidity is vital in wildfire prevention and control (Agrawal et al., 2023; Si et al., 2022). When relative humidity is high, the surrounding environment contains abundant moisture, keeping plants and soil moist. In these conditions, plants dry out less frequently, which decreases the chance of fire occurrence. Moreover, high humidity fosters a moist and green vegetation cover, reducing the availability of combustible fuels for fires (Jain et al., 2022). This research utilized data from the FLDAS model as a source for humidity information.

- **Wind speed:** Wind speed is critical in identifying wildfire-prone areas (Kong, 2024; Li et al., 2021). It directly influences fire spread by accelerating the rate at which fire travels. High wind speeds can also disperse fire embers, expanding the affected area and increasing wildfire risk (Li et al., 2021). This study utilized wind data from the FLDAS model to analyze mentioned effects.

- **Evapotranspiration:** Evapotranspiration is the process by which water is released from a vegetated area through evaporation from the soil surface and transpiration from plants (Pereira et al., 1999). A high evapotranspiration rate indicates a high moisture level in the vegetated area, as water continuously evaporates from the soil and plants, keeping the area moist and reducing fire risk. This process saturates with moisture. Conversely, low evapotranspiration means limited moisture, reducing evaporation from the soil and plants (Zhu et al., 2024). This situation increases the fire risk and the speed at which fires can spread (Bond-Lamberty et al., 2009). This study utilized the MODIS evapotranspiration product, which had an 8-day temporal resolution.

- **Land surface temperature:** Land Surface Temperature (LST) serves as an indicator of drought conditions and wildfire risk (Çolak & Sunar, 2023; Sharma & Dhakal, 2021). Higher LST often signifies dryness and water scarcity within vegetation. Under these circumstances, the earth's surface and vegetation tend to dry out, creating favorable conditions for fire occurrence. Moreover, increased LST accelerates water evaporation from soil and plants, further reducing vegetation humidity. Therefore, higher LST values typically correlate with heightened wildfire risks (Guangmeng & Mei, 2004). This study utilized daily LST data derived from the MODIS sensor.

- **Soil moisture:** Soil moisture refers to the water content present in the soil and serves as a critical indicator for assessing wildfire conditions (Colak & Sunar, 2020; Fadaei et al., 2022). High soil moisture levels indicate sufficient water content in the soil and vegetation. This saturation helps prevent the accumulation of dry vegetation, which is highly susceptible to ignition. According to Table 1, this study utilized the soil moisture index derived from the SMAP sensor, which provided data at a resolution of 10 km.

- **Land cover type:** It is necessary to consider the type of land cover when identifying wildfire-prone areas. Different land cover types can reduce or increase the possibility of fire occurrence in specific locations (Cano-Crespo et al., 2023; Novo et al., 2020). Identifying flammable materials, such as dry leaves or vegetation, is essential for assessing fire risk. The composition of plant species in the forest significantly influences the identification of fire-prone spots; certain species, such as spruce, cypress, and pine, contain highly flammable materials that can facilitate rapid fire spread if ignited. Moreover, the density of vegetation per unit area also plays a significant role in fire risk assessment, as high densities of dry and flammable plants increase the likelihood of rapid-fire spread (Koetz et al., 2008). In this study, the land cover variable, crucial for identifying wildfire-prone areas, was encoded using label encoding. As illustrated in Figure 3, which depicts the types of land cover in the study area, the spatial distribution of vegetation plays a crucial role in understanding its potential influence on fire risk.

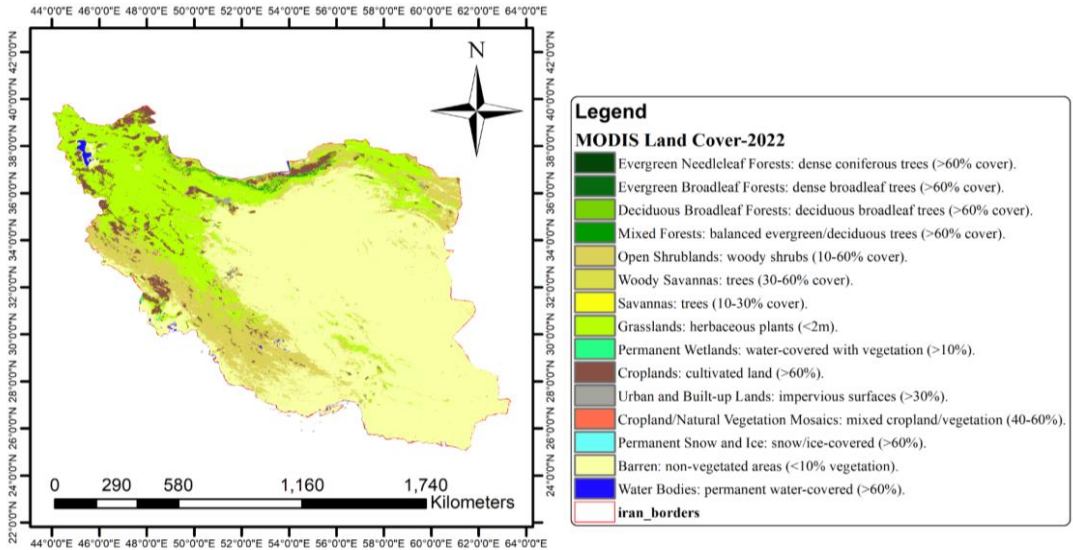

Figure 3 : Spatial Distribution of land cover type in the study area.

### 3.1.4    Anthropogenic parameters

This section introduces human-related factors influencing the occurrence of wildfires. Human activities, such as discarding cigarettes, unauthorized burning, and improper waste disposal, significantly increase fire risks. Moreover, poor safety protocols and inadequate management in industrial areas can trigger fires that may spread to nearby forests (Berčák et al., 2024; El Mazi et al., 2024; Kolanek et al., 2021; Pragya et al., 2023; Sikuzani et al., 2023).

The human-related factors were produced using spatial layers sourced from the Overpass Turbo platform. Vector data in point and line file formats were invoked. Then, Euclidean distances and density metrics were calculated to evaluate the impact of human activities and infrastructure on wildfire occurrence. Point data, such as the locations of industrial sites and population centers, enabled the calculation of proximity to wildfire-prone regions, while line data, including road networks and power transmission lines, determined distances from high-risk fire zones. The proximity of human-related factors industrial points, roads, population centers, population density, and power lines plays a pivotal role in identifying areas vulnerable to wildfires.

Table 1 distinguishes between static and dynamic parameters utilized in this study. Static parameters include anthropogenic and topographical factors, while climatic and vegetation indices are classified as dynamic variables.

## 3.2    A framework for wildfire susceptibility mapping

This study integrates topological, meteorological, remote sensing data, and spatial information layers to identify wildfire-prone areas in Iran. It also assesses the influence of each parameter under various wildfire occurrence scenarios. Figure 6 provides an overview of the research methodology.

After preprocessing the data, several machine learning algorithms were employed to model the relationship between input parameters and wildfire occurrence. The study then compared the performance of various machine learning models under different scenarios, focusing on selecting training and test samples. Finally, the best model generated a wildfire susceptibility map for the study area.

## 3.3    Data preprocessing

### 3.3.1    Calculate anomaly

As shown in Table 1, investigating and identifying wildfire-prone points requires analyzing static and dynamic parameters over time. Static parameters remain constant, while dynamic parameters change. To better understand the impact of dynamic parameters, we focused on their anomalies instead of their raw values. Anomalies unusual fluctuations are deviations from expected values, specifically the average of the relevant parameters. Detecting these anomalies can significantly improve the identification of wildfire-prone areas.

Equation 1 illustrates the calculation of anomalies, where $Feature_i$ represents the value of each variable at a specific time, and $Feature_{mean}$ is the average value of that feature over the study period.

$$\text{Anomaly} = Feature_i - Feature_{mean} \qquad (1)$$

### 3.3.2    Missing value interpolation

Some parameters derived from satellite observations can be impacted by issues like cloud cover, leading to pixels without values. This absence of data complicates analysis and modeling. To address this, two methods K-nearest neighbors (KNN) and K-means clustering—were used to fill in missing pixels based on nearby data.

In the KNN method, the k nearest neighbors with available values are identified for each pixel without a value, and their average is used to estimate the missing pixel. Meanwhile, in the K-means method, the dataset is divided into clusters, and the missing values are replaced with the average values of their respective clusters. The neighborhood radius in both methods is adjustable, depending on data conditions and the number of nearby pixels lacking values.

Figure 4 and Figure 5 provide an overview of the parameters employed in this research. Dynamic parameters are visualized for the year 2022. The DEM shows that Iran's mountainous regions, particularly the Zagros and Alborz regions, are more prone to wildfires due to significant elevation changes and steep slopes. Steep slopes facilitate the rapid spread of fires, while south-facing slopes are drier and warmer, increasing the likelihood of wildfires. In forested regions of northern and central Iran, vegetation indices such as EVI and NDVI, which measure vegetation health and density, play a significant role in wildfire analysis. However, the influence of human factors cannot be overlooked. The presence of industrial sites, population centers, and proximity to road networks and power transmission lines significantly heighten the risk of wildfires, underscoring the need for human intervention and policy changes. Environmental conditions, particularly high temperatures and low humidity in southern Iran, significantly elevate fire risks. High wind speeds in southwestern Iran further accelerate the fire spread. These findings underscore the urgent need to address climate change and its effects. Finally, dry land cover and low soil moisture in southern provinces like Fars, combined with high evapotranspiration and low precipitation, create favorable conditions for wildfires.

### 3.3.3 Normalized Mutual Information for feature analysis

Mutual information (MI) quantifies the degree of mutual dependence between two random variables. It measures how much knowing the value of one random variable reveals information about the value of the other. This concept is intrinsically related to the entropy of a random variable, which represents the amount of information it contains. Equation 2 demonstrates how to calculate mutual information, where $X$ and $Y$ are two random variables. $P(x, y)$ denotes the joint probability of the occurrence of $x$ and $y$. At the same time, $P(x)$ and $P(y)$ represent the individual probabilities of events $x$ and $y$, respectively.

$$MI\ (X, Y) = \sum_{x \in X} \sum_{y \in Y} P(x, y) log \frac{P(x,y)}{P(x)P(Y)} \qquad (2)$$

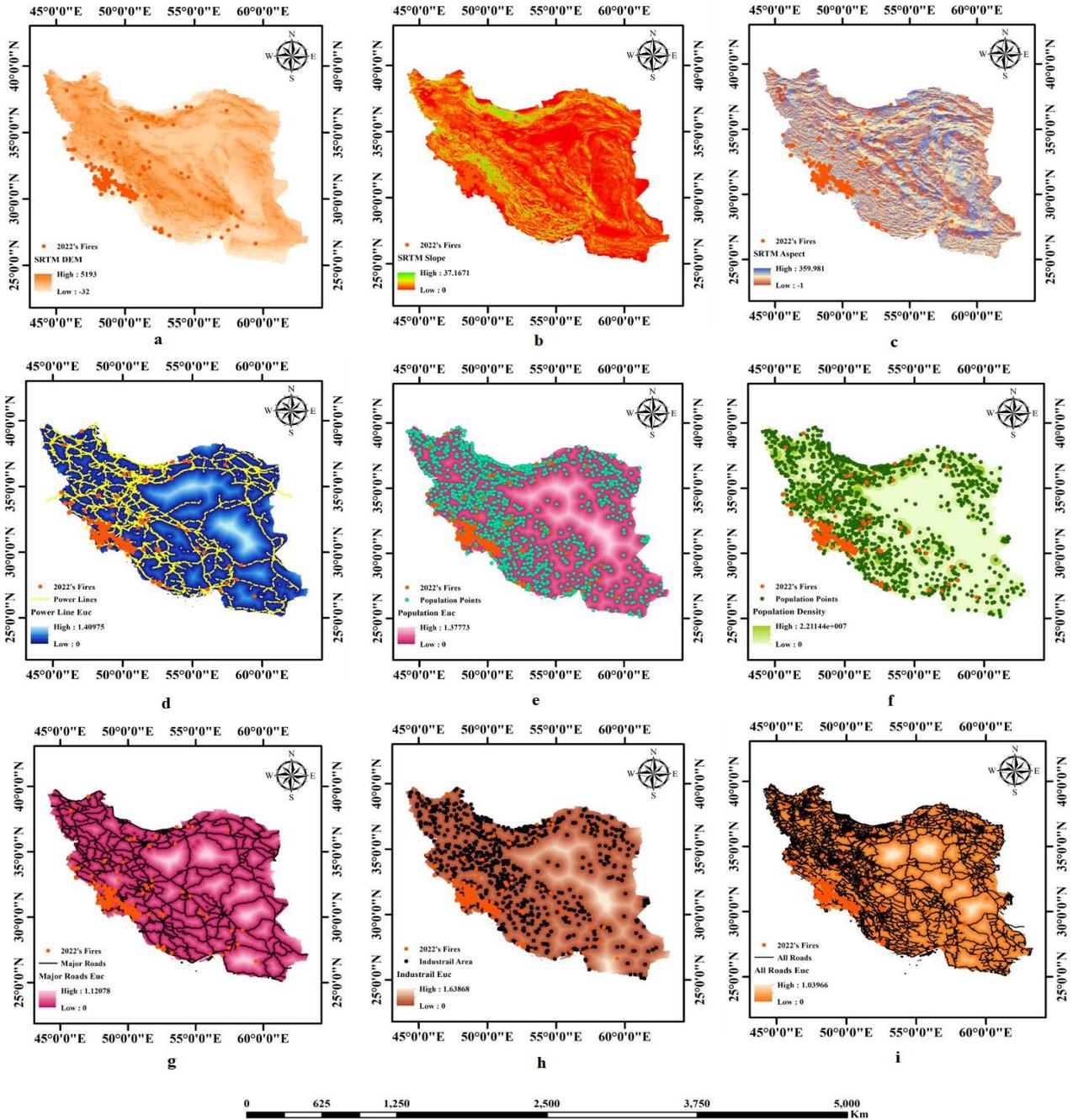

Figure 4: Display of the static input features:  a) DEM, b) Slope, c) Aspect, d) Power line EUC, e) Population EUC, f) Population Density, g) Major Road EUC, h) Industrial EUC, i) All Road EUC.

Conversely, features with lower NMI scores may offer limited predictive insights individually, but combined with others, they can still play a critical role in building comprehensive models (Chopade et al., 2024; Jerdee et al., 2023; Theng & Bhoyar, 2024).  By calculating the NMI for each feature, researchers can rank them based on relevance, ensuring that the most influential features are prioritized for constructing accurate and robust fire risk models.

In the domain of feature importance, Normalized Mutual Information (NMI) extends mutual information (MI) by normalizing the measure, allowing for standardized comparisons across different features. NMI is a powerful tool in determining the contribution of each feature to predicting or explaining the target variable. A higher NMI score indicates a stronger relationship between the feature and the target, suggesting that it provides significant information relevant to fire risk prediction.

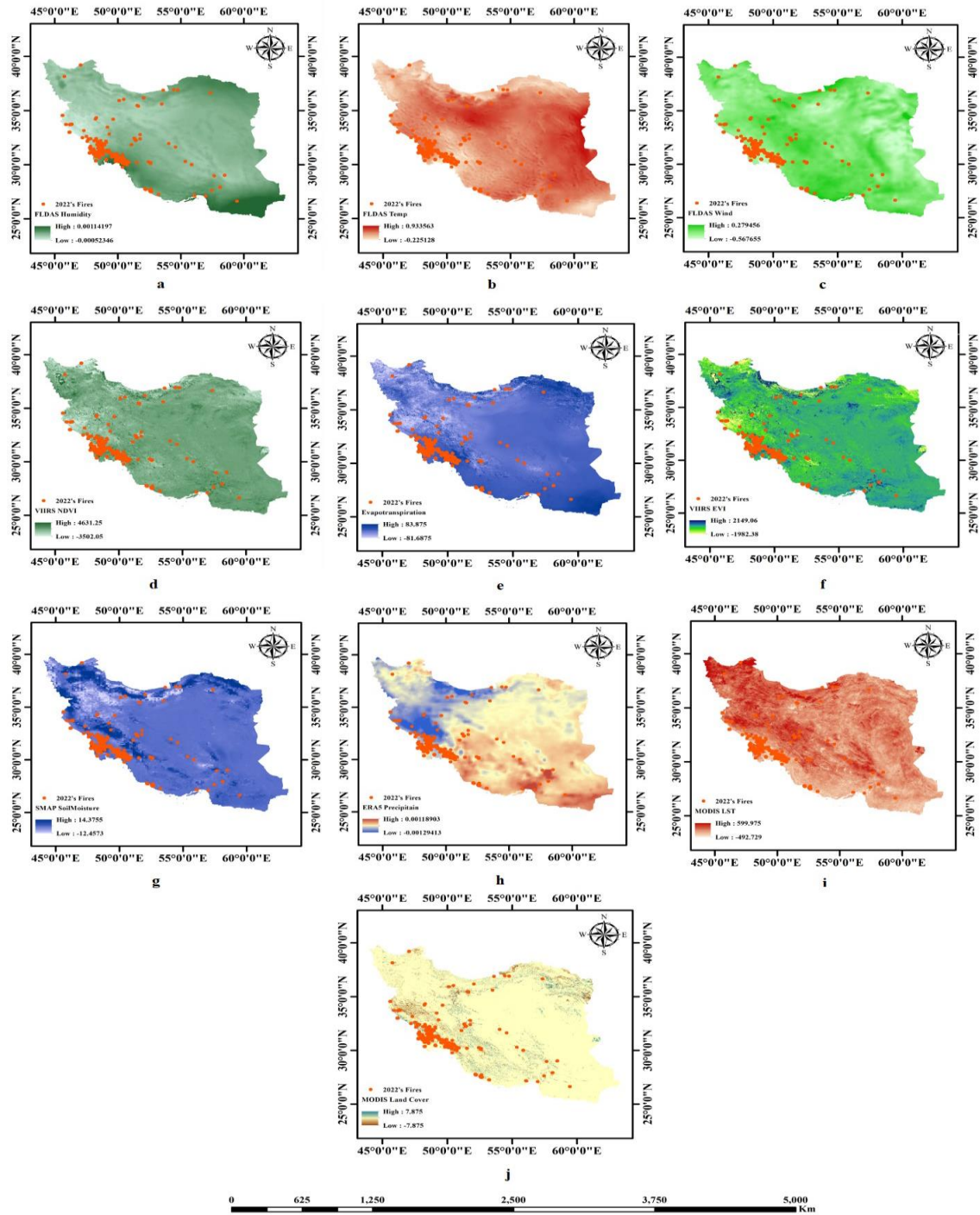

Figure 5 : Display of the dynamic input features for 2022: a) Humidity, b) Temperature, c) Wind, d) NDVI, e) Evapotranspiration, f) EVI, g) Soil Moisture, h) Precipitation, i) LST, j) Land Cover.

### 3.3.4    Variance Inflation Factor for feature analysis

Variance Inflation Factor (VIF) tests how the behavior of an independent variable changes due to its correlation with other independent variables, assessing how strongly these variables are correlated and how this affects model fit. High dependence between two variables means that even small changes in the input data structure or the regression model can lead to significant and sometimes erratic changes in the variable coefficients (Akinwande et al., 2015; Juarto, 2023). Equation 4 illustrates how to calculate VIF, defined as:

$$VIF_i = \frac{1}{1 - R_i^2} \qquad (3)$$

where $R$ is the correlation between the independent variable and other variables. VIF also helps identify which variables should be removed to enhance the model's reliability and prevent an undesirable increase in the variance of the coefficients (Frost, 2017). Using VIF to detect multicollinearity with categorical variables presents specific limitations and challenges. It is necessary to consider alternative approaches or modifications to address this issue. Thus, Label Encoding was employed to convert categorical variables into numerical values. This approach is widely used for transforming categorical variables into a format suitable for statistical analyses like VIF. In this method, each category or class of the categorical variable is assigned a unique integer value, effectively enabling the use of the data in multivariate analyses. The encoding process was structured to reflect each class's priority and potential influence on wildfire occurrences, ensuring the optimal application of Label Encoding in the analysis.

## 3.4    Machine learning based identification of wildfire susceptibility

In this research, several techniques were also applied to evaluate the significance of different input parameters, including anthropogenic and climatic factors, in determining wildfire-prone areas. Each phase of the research will be explained in detail in the following sections.

This research employed a comprehensive approach, utilizing a range of machine learning algorithms such as Random Forest (RF), Extreme Gradient Boosting (XGBoost), KNN, Support Vector Machine (SVM), Decision Tree (DT), and Gradient Boosting Tree (GBT) to identify wildfire-prone areas.

Independent variables, including topographic features, vegetation indices, climatic parameters, and anthropogenic factors, were meticulously selected, and their values were extracted from locations where wildfires had occurred.

Prior to analysis, a high-precision 25-meter spatial resolution forest mask was employed to accurately differentiate forested areas from non-forested areas. This mask was instrumental in defining various research scenarios, ensuring that training and test points were drawn from the most relevant areas.

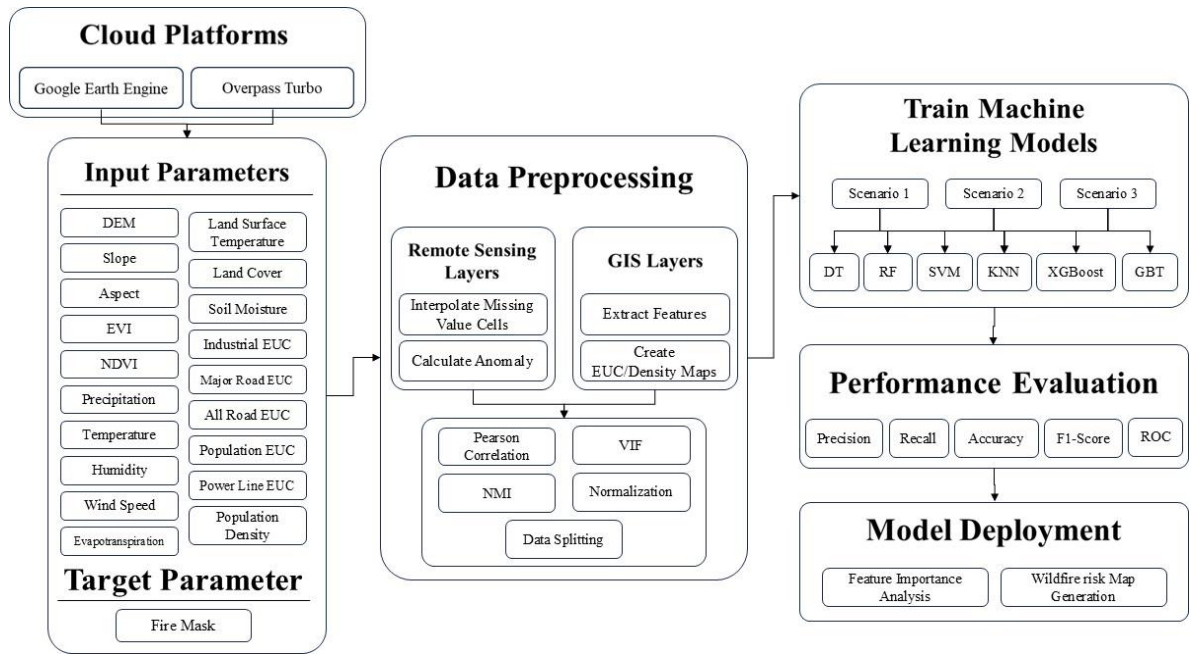

Figure 6: The framework implemented for wildfire susceptibility assessment

To identify locations of past wildfires, a fire mask with a 1000-meter spatial resolution, derived from MODIS observations, was used. This fire mask was employed to select fire-occurrence points. In this mask, wildfire locations have been classified into nine classes based on wildfire probability. For greater reliability, only points categorized in Class 9, representing the highest probability of wildfire, were selected.

Thus, fire-occurrence points (label 1) and non-fire points (label 0) were carefully selected. This distinction was crucial, enabling the machine learning models to differentiate between the two categories accurately.

### 3.4.1 Machine learning models

As a fundamental model, the Decision Tree (DT) algorithm was used for wildfire susceptibility analysis. It follows a hierarchical structure based on sequential decisions, making it a foundational method in this analysis. The tree generates nodes and branches by splitting data according to attributes and values, enabling complex decision-making through input features (Costa & Pedreira, 2023; Priyanka & Kumar, 2020).

Random Forest (RF) enhances the decision tree approach by constructing multiple decision trees independently through random sampling of the dataset. The final prediction is achieved by averaging the outputs of all the trees, which improves the model's robustness and significantly reduces the risk of overfitting compared to a single decision tree (Parmar et al., 2019; Tariq et al., 2023). This ensemble method leverages the diversity of the individual trees, allowing for more stable and accurate predictions across various datasets.

Gradient Boosting Trees (GBT) construct a robust model by sequentially combining several weak decision trees. Each weak model is trained on the dataset, with its prediction calculated errors. An optimization algorithm, often gradient descent, is employed to develop the next model, concentrating on minimizing the errors of the preceding model. This iterative process involves generating a new model that improves predictions in the feature space by determining the optimal update based on the gradient (partial derivative) of prior prediction errors (Bhagwat & Shankar, 2019).

XGBoost is a widely used decision tree-based machine learning algorithm that enhances the gradient boosting method. At each iteration, a weak decision tree is created to correct the errors made by the previous model. These trees are subsequently combined sequentially to produce the final model. XGBoost is notably optimized and utilizes parallel processing, resulting in superior speed compared to other similar algorithms (Parmar et al., 2019; Ren et al., 2024).

This study also examines the performance of traditional classification models, including K-Nearest Neighbors (KNN) and Support Vector Machines (SVM), alongside decision tree-based methods. KNN, a straightforward algorithm, operates on the principle of nearest neighbors by calculating the distance between an unknown sample and all training samples. It selects the K closest samples based on this distance, assigning the predominant label among these neighbors as the class of the unknown sample. This approach is valued for its computational efficiency due to its simplicity.

Similarly, SVM is employed for classification and regression tasks, separating distinct classes by defining a hyperplane in the feature space. The primary objective of SVM is to find the optimal hyperplane that maximizes the margin between classes, ensuring the most significant possible distance between the nearest points of different classes (Mountrakis et al., 2011; Samaniego & Schulz, 2009).

### 3.4.2 Model setup, hyperparameter tuning and evaluation metrics

This research used data from 2015 to 2021 to train the machine learning models, while data from 2022 served as the test set. Hyperparameter tuning was executed using the common K-fold cross-validation method with K=5. Table 2provides the hyperparameters for the various models employed in this study.

Table 2: The hyperparameters tuned for the various classifiers.

| Models | Hyper-parameters |
|--------|------------------|
| DT | splitter = best |
| RF | n estimators = 100, min samples split = 2, min samples leaf = 1 |
| SVM | gamma = auto, probability = True, C = 1000, kernel = rbf |
| KNN | n neighbors = 5, weights = distance |
| XGBoost | n estimators = 200, max depth = 5, col sample by tree = 0.9, learning rate = 0.2, subsample = 0.8 |
| GBT | n estimators = 200, max depth = 5, learning rate = 0.2 |

This study used the F1 score, Recall, Precision, and Receiver Operating Characteristic (ROC) metrics for performance assessment of classifiers. Precision indicates the proportion of true positive fire detections among all detected fires, demonstrating how many of the detections were accurate. A high Recall means that the algorithm successfully identified most actual fire occurrences, resulting in few false negatives. The F1 score combines Precision and Recall, offering a balanced assessment of the algorithm's fire detection accuracy. A higher F1 score reflects better overall performance.

The ROC metric evaluates model detection power analytically. The ROC graph illustrates model performance as the decision threshold varies, with the horizontal axis representing the False Positive Rate (FPR) and the vertical axis representing the True Positive Rate (TPR). A model performs better if the ROC curve is closer to the top left corner and the larger area under the curve (AUC). The AUC indicates overall model performance, ranging from 0 to 1. An AUC of 1 signifies excellent performance, accurately identifying all wildfire instances without misclassifying non-wildfire points. Conversely, an AUC of 0.5 indicates random performance (Kalantar et al., 2020).

### 3.4.3 Various scenarios for wildfire risk assessment

This study evaluates parameters influencing the identification of wildfire-prone areas, utilizing models such as KNN, DT, RF, SVM, GBT, and XGBoost, all constructed with optimized parameters. A crucial component of these experiments is the meticulous selection of training and testing samples to establish diverse scenarios for model evaluation. Three distinct scenarios were designed to assess wildfire risk modeling. The details of each scenario will be discussed in the following.

- **Scenario 1**

In this scenario, both wildfire occurrence and non-occurrence samples were exclusively selected from forested areas in Iran. To minimize spatial bias, the selection of non-fire samples was carefully designed. Specifically, non-fire points were distributed geographically across the seven-year study period, ensuring that the same locations were not repeatedly used as non-fire samples in different years. This approach prevented the models from overfitting to the specific characteristics of particular locations, such as soil type, vegetation cover, or microclimatic conditions. By incorporating geographic and temporal variability, the scenario tested the models' ability to accurately detect wildfires in diverse forested environments. Additionally, this method acknowledged the dynamic nature of forest ecosystems, where changing conditions over time could influence wildfire susceptibility

- **Scenario 2**

Similar to Scenario 1, samples were selected from forested areas, but a minimum distance of 25 km was maintained between training and testing points for both classes. This spatial separation was applied between the training and testing samples to evaluate the models' ability to generalize to new areas. This distance ensured that the training data did not include features similar to those in the testing samples, thus preventing data leakage. Data leakage occurs when test data is too similar to training data, leading to inflated performance metrics that do not accurately reflect the model's true predictive power. The spatial separation also emphasized the practical application of the models in real-world scenarios, such as predicting fire risks in regions where detailed historical data is unavailable. By simulating these conditions, the scenario demonstrated the models' capacity to apply knowledge from known areas to new, potentially more challenging environments. Focusing on forested areas ensured the relevance of the results for wildfire management in similar ecosystems.

- **Scenario 3**

This scenario expanded the analysis scope by including forested and non-forested regions. Unlike previous scenarios focused solely on forests, this approach incorporated diverse landscapes, such as shrublands, grasslands, and sparsely vegetated areas. This broader dataset aimed to capture the complexity of wildfire dynamics across diverse land cover types in Iran. Wildfires are not confined to dense forests; they can occur in transitional zones, parklands, or areas with scattered vegetation. By removing restrictions on sample selection, this scenario assessed the models' ability to detect wildfire risk in non-traditional environments. The comprehensive dataset also provided insights into how environmental factors, such as soil moisture and proximity to human settlements, influence fire occurrence. Including non-forested regions enhanced the robustness of the analysis and ensured the models' applicability for fire risk assessments across a wide range of ecological contexts.

## 3.5   Features importance analysis and wildfire susceptibility mapping

The primary objective of this study is to assess how human and climatic factors contribute to wildfire susceptibility. By focusing on these factors, different techniques will be employed to evaluate feature importance, elucidating the relative impact of each parameter on wildfire risk. This approach will provide detailed information on the significance of each feature and its relationship with the target variable, allowing for identifying the most critical parameters in the wildfire-prone spot detection model. Additionally, it helps recognize features with a lesser individual impact but a more meaningful influence when considered in conjunction with other features.

Machine learning methods such as XGBoost, RF, and GBT can be utilized to assess the importance of features. These methods evaluate the importance of each feature based on its impact on model error reduction (Theng & Bhoyar, 2024). In XGBoost, feature importance is calculated by measuring the reduction in error at various tree splits. Random Forest determines feature importance by assessing how features reduce uncertainty at decision tree nodes. Gradient Boosting Trees measure feature importance based on each feature's contribution to enhancing the performance of previous models. Due to their ability to capture complex data relationships and accurately evaluate feature relevance, these methods are effective tools for modeling susceptibility to forest fires.

After training, the machine learning models served as classifiers, generating probability values for each pixel and indicating the likelihood of fire occurrence. These raw probability values were first used to create maps before being rounded to show the presence or absence of fire. To categorize the risk levels into five distinct categories from very low to very high risk a data-driven method was employed to differentiate among these values, ensuring a clear distinction between the various risk levels represented in the maps.

# 4    Results

## 4.1    Results of feature analysis in wildfire susceptibility modeling

Figure 7 presents a correlation analysis of the prepared dataset, illustrating the relationships among various parameters. Each cell's numerical value indicates the linear correlation, while the shading intensity reflects the strength of this correlation. This analysis provides insights into the significance of each parameter's impact on others and the target parameter.

One of the most critical steps in data analysis is to investigate the linear dependence between input features. This step is crucial as it ensures the validity and reliability of our results. While uncorrelated features can be analyzed independently, it is crucial to account for dependencies when they exist. This highlights the importance of considering these relationships in our analysis process.

Notable correlations identified include:

- 0.64 between the distance from main roads and power transmission lines

- 0.68 between the distance from population centers and industrial areas

- 0.61 between the distance from main roads and secondary roads

- 0.89 between NDVI and EVI indices

- 0.70 between humidity and precipitation

These correlations stem from various factors. For instance, the correlation between the distance from main roads and power transmission lines is often due to similar infrastructure designs. The high correlation between population centers and industrial areas reflects the concentration of industries near densely populated regions. NDVI and EVI assess vegetation cover, explaining their high correlation, while humidity and precipitation are naturally linked as precipitation provides environmental moisture. Overall, the observed linear correlations, while not excessively high, are significant, suggesting a well-considered selection of features that likely yield valuable modeling information and emphasizing their potential in data analysis.

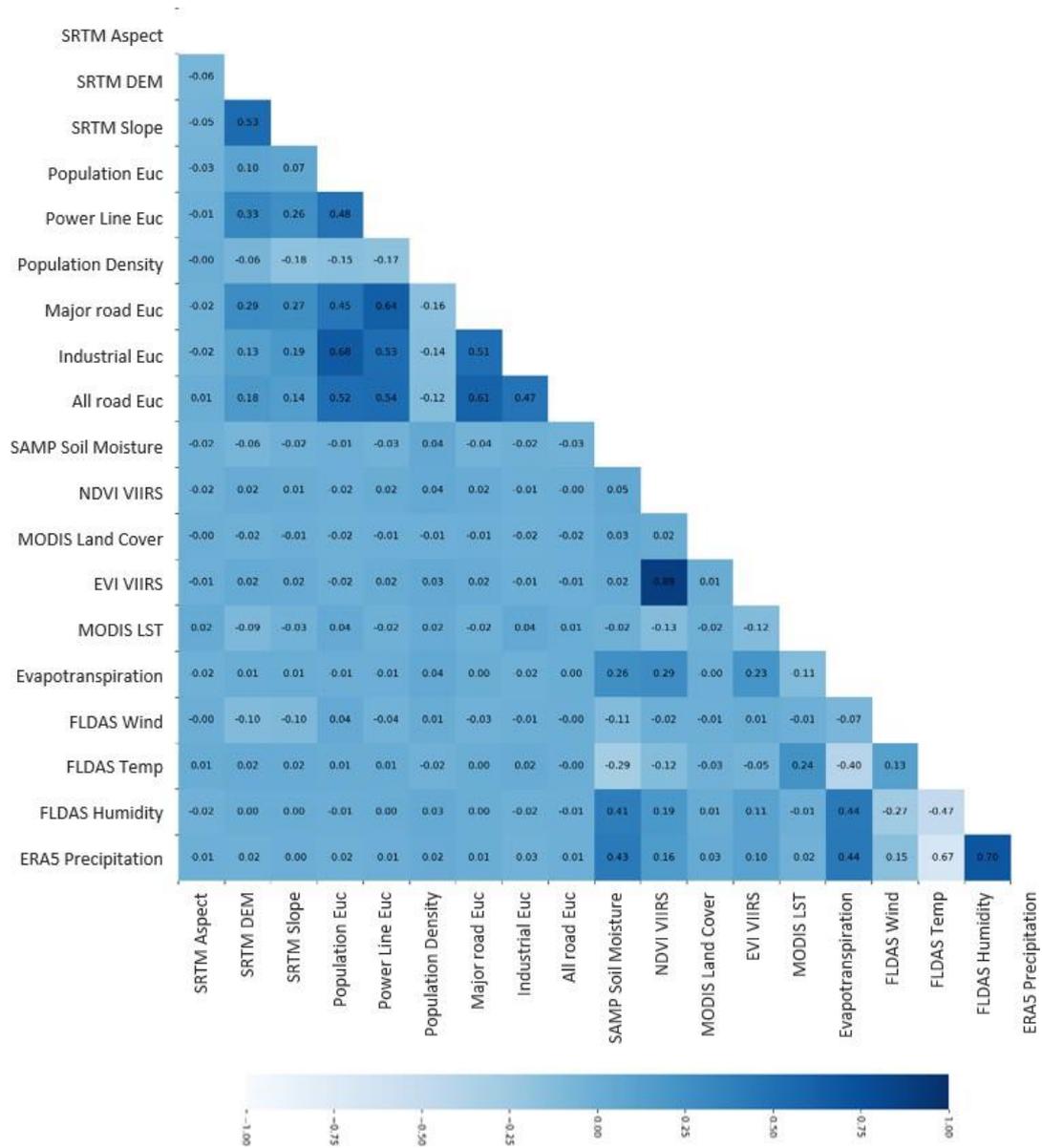

Figure 7 : The correlation heat map of input parameters employed for wildfire susceptibility modeling

While correlation analysis is a simple and widely used method, it has notable limitations. One significant drawback is its ability to capture only linear relationships between features, making it unsuitable for identifying complex, nonlinear interactions. Additionally, correlation analysis focuses solely on the associations between input features and does not assess the relationship between these features and the target variable. Consequently, more advanced techniques are required to effectively capture nonlinear and intricate dependencies between features and the target variable.

While linear correlation analysis indicates the relationship between input parameters, it does not account for the impact of the target variable. Therefore, advanced methods should be employed to investigate the influence of various parameters on wildfire occurrence. In this regard, the NMI method was used to evaluate the impact of different parameters in identifying wildfire-prone areas. Figure 8 illustrates the effect of each parameter on the target variable according to this index. The parameter

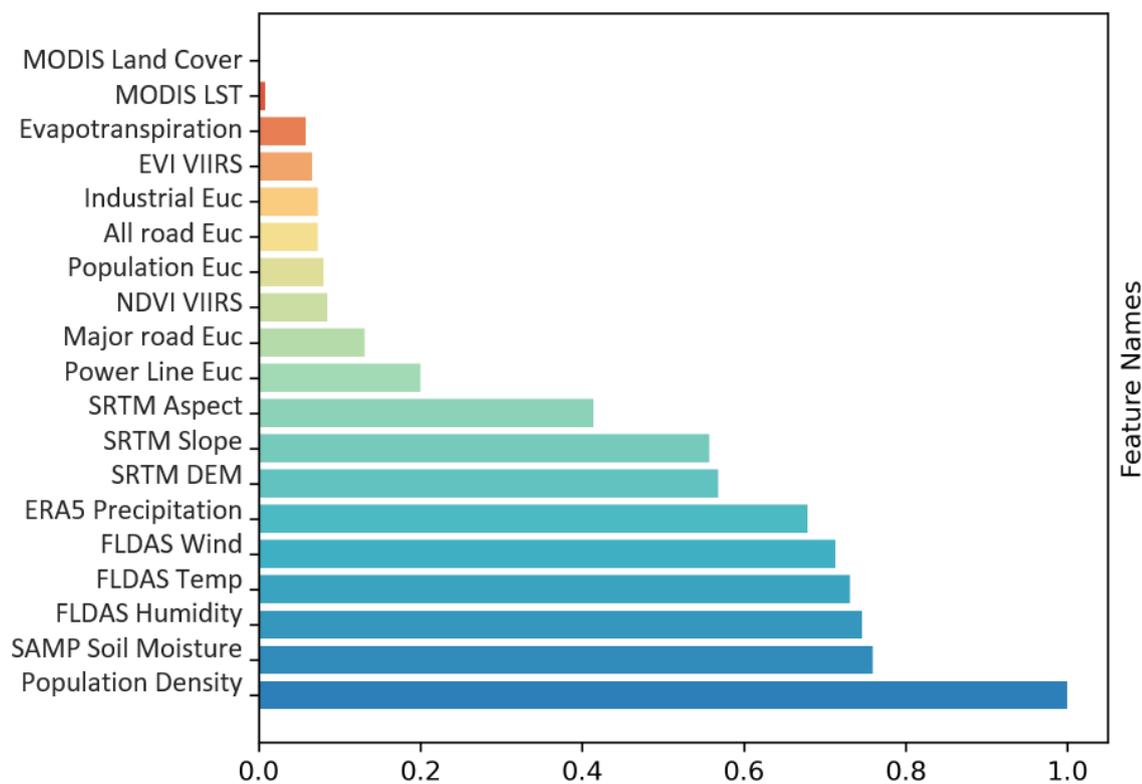

Figure 8: Features importance analysis for wildfire susceptibility assessment using NMI. Blue indicates a higher reliance on the target, while red represents a lower dependence on the target for each input parameter.

Land Cover has the lowest dependency (NMI value of zero), suggesting that it has minimal impact on identifying wildfire-prone areas.

In contrast, Population Density, with an NMI index value of 1, has the most significant effect on the target variable and is the most significant feature in identifying wildfire-prone areas. Other important parameters in this analysis include soil moisture, humidity, temperature, wind, and precipitation, each with an NMI above 0.68. Parameters such as surface temperature, evapotranspiration, EVI, distance from industrial areas, and distance from roads were found to have less impact.

In this study, the importance of features in assessing high-risk forest fire areas was also evaluated using the VIF test (Figure 9). The Variance Inflation Factor (VIF) identifies multicollinearity issues by highlighting undesirable linear relationships among variables. High correlation between variables can lead to redundant information, inflating the standard errors of regression coefficients and reducing their reliability. This strong correlation may also result in unstable coefficient estimates, where minor changes in the data or model can cause significant variations in the coefficients. A high VIF value, typically above 10, indicates that a variable is highly correlated with other variables, which can negatively impact

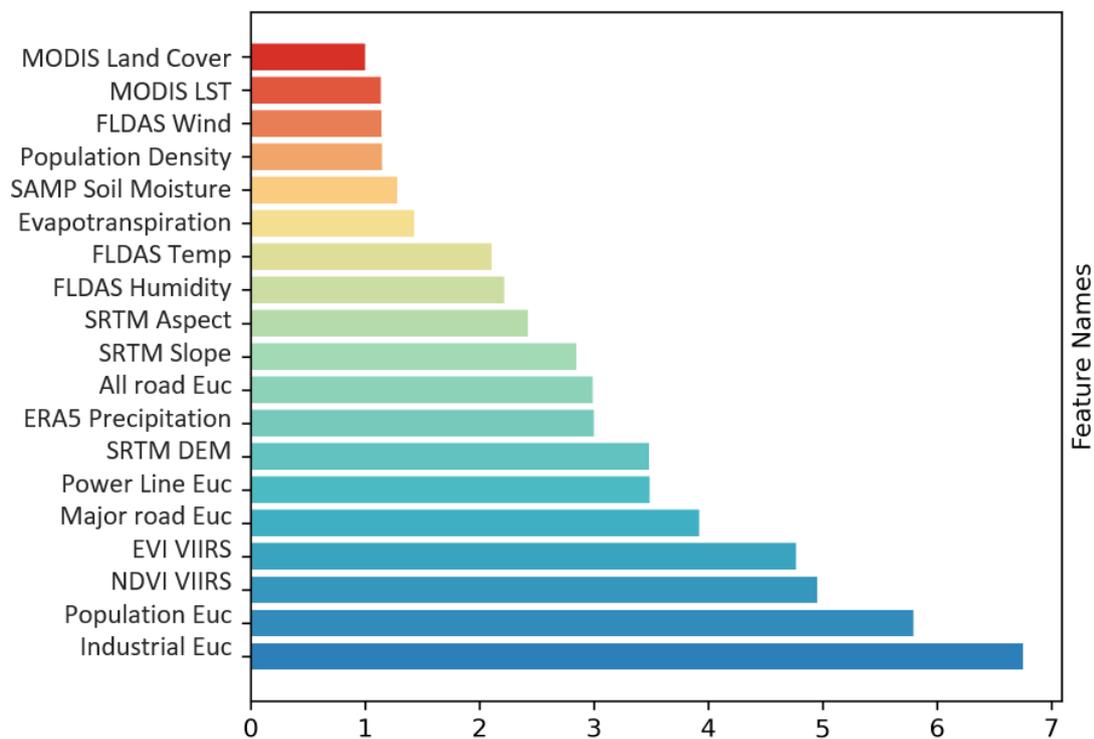

Figure 9: Features importance analysis for wildfire susceptibility assessment using VIF. By transition from red to blue, the VIF index becomes increasingly unfavorable.

the model's accuracy and stability.

From the results, it has been determined that for all input features, the VIF values are less than 10, indicating balanced correlations. The least effective VIF values were observed for distance from industrial areas at 6.76, distance from population centers at 5.79, NDVI index at 4.96, and EVI index at 4.77. This suggests that these features have lower levels of linear correlation with other variables in the model.

## 4.2 Result of modeling wildfire susceptibility

### 4.2.1 Scenario 1

In Scenario 1, two distinct types of samples were utilized: points located within forest areas where fires have occurred and points within forest areas that have remained fire-free. For each year during the study period, the locations of these points were treated as separate observations. This data structure enabled the training of various parameters and models to analyze fire occurrence patterns effectively.

In the initial analysis, the importance of input parameters was evaluated using RF, XGBoost, and GBT algorithms, as illustrated in Figure 10. The findings indicate that all three algorithms identify slope, elevation, and distance from populated areas as the most critical parameters for detecting and predicting wildfires in forested regions. These parameters significantly influence the model's decision-making and are thus considered critical factors in wildfire detection.

Conversely, LST and land cover type are deemed less necessary in the model's decision-making process. All three algorithms consistently show that these parameters contribute less to wildfire detection than the more influential ones mentioned earlier. Additionally, the consistent ranking of parameter importance across RF, XGBoost, and GBT reinforces the reliability of these findings, suggesting that the identified importance of parameters is valid across different modeling approaches.

In the second step, the performance of various machine learning models was assessed using the ROC plot. As illustrated in Figure 11, the RF, XGBoost, and GBT models exhibited the highest performance, achieving an AUC of 98%. In comparison, the KNN, DT, and SVM models recorded AUC values of 94%, 87%, and 84%, respectively.

Table 4 presents the performance assessment of the implemented models based on various metrics. As shown in Table 3, the RF model outperformed all other models across four key metrics: F1-score, recall, precision, and accuracy. The XGBoost and GBT models also performed well, securing second and third places in all accuracy metrics, making them strong candidates for addressing this problem. In contrast, the SVM and DT, and KNN models demonstrated lower accuracy than the others.

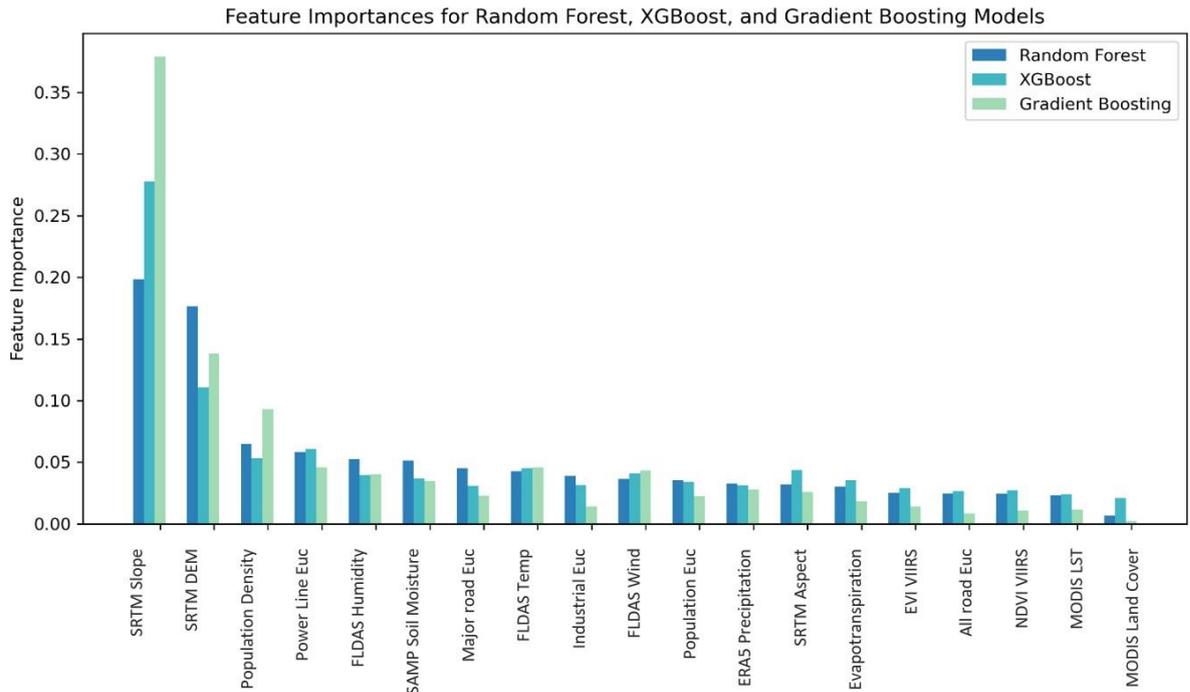

Figure 10: Feature importance analysis for wildfire susceptibility assessment using RF, XGBoost, GBT models based on scenario 1.

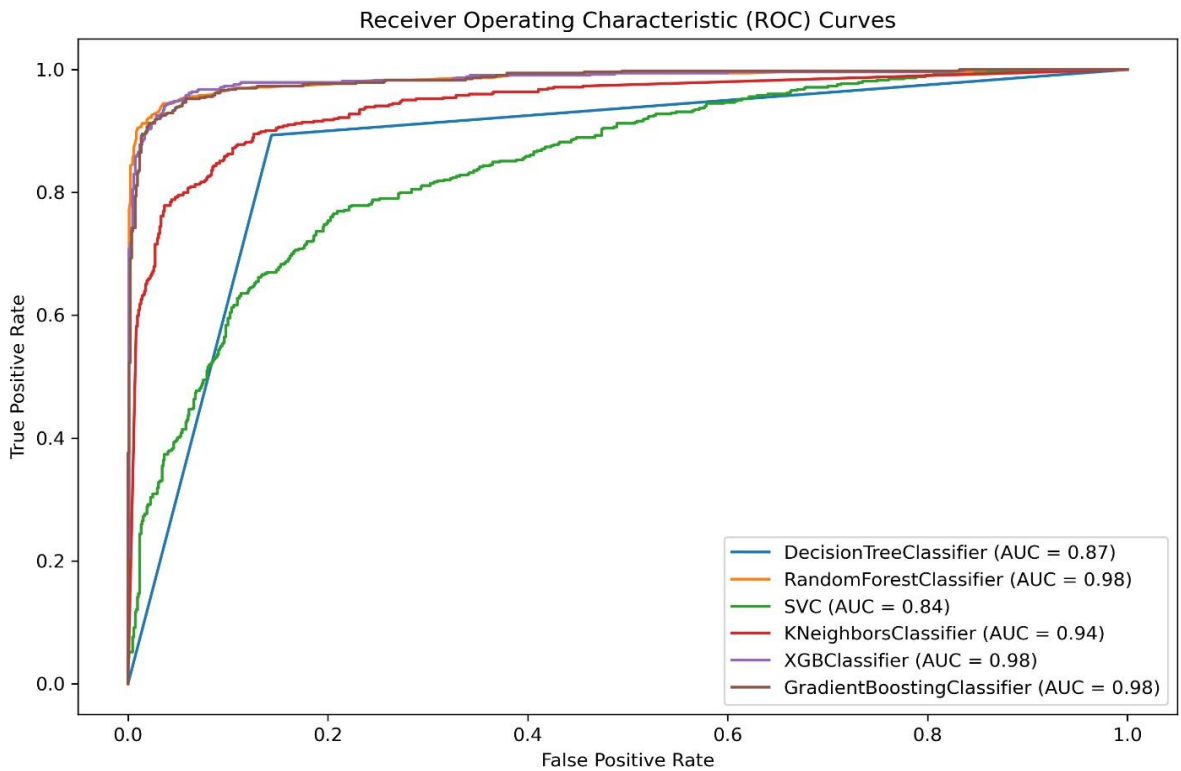

Figure 11: ROC diagram of different algorithms for scenario 1.

Table 3: The results of different classification algorithms based on F1-score, recall, precision and accuracy metrics for scenario 1.

| Metrics/Models | GBT | XGBoost | KNN | SVM | RF | DT |
|---|---|---|---|---|---|---|
| **Accuracy** | 0.946 | 0.951 | 0.880 | 0.787 | 0.956 | 0.870 |
| **Precision** | 0.946 | 0.951 | 0.884 | 0.786 | 0.956 | 0.877 |
| **Recall** | 0.946 | 0.951 | 0.880 | 0.787 | 0.956 | 0.870 |
| **F1-score** | 0.946 | 0.951 | 0.881 | 0.782 | 0.955 | 0.872 |

### 4.2.2 Scenario 2

In Scenario 2, the test dataset was selected to ensure that the training samples were not located within a few kilometers of the test samples. This approach ensures that the model is not familiar with these specific locations. Consequently, the test samples were chosen to have no significant spatial dependency or neighborhood relation with the training data.

Like Scenario 1, the RF, GBT, and XGBoost algorithms were employed to assess the importance of input parameters in identifying wildfire-prone areas. The results indicate that LST and land cover type have the least impact on wildfire susceptibility. In contrast, elevation, slope, and distance from power lines significantly influence identifying wildfire-prone areas.

Comparing Figure 10 and Figure 11, it is evident that while there are minor variations, the importance of parameters between Scenarios 1 and 2 remains relatively stable in forest environments.

Figure 13 illustrates the performance comparison of various models using the ROC plot. The RF and XGBoost algorithms perform best, achieving an AUC of 94%. This high AUC indicates a substantial likelihood of correctly identifying wildfire-affected and fire-free areas. The GBT algorithm follows closely, ranking second with an AUC of 93%, showcasing its effectiveness in distinguishing between wildfire-prone and fire-free regions.

Next, the KNN and SVM algorithms have 86% and 68% of AUCs, respectively, indicating moderate performance in this context. In contrast, the DT algorithm displays the lowest performance with an AUC of 61%, reflecting its lower accuracy and reduced effectiveness in classifying wildfire-prone areas.

Table 4 presents the performance of various models based on F1-score, recall, precision, and accuracy for Scenario 2. The KNN algorithm achieves the highest F1-score of 0.687, indicating its effectiveness in accurately identifying categories. Following KNN, the RF and XGBoost algorithms yield F1-scores of 0.443 and 0.587, respectively. In contrast, the Gradient GBT, SVM, and DT algorithms show lower performances with F1-scores of 0.515, 0.311, and 0.521, respectively.

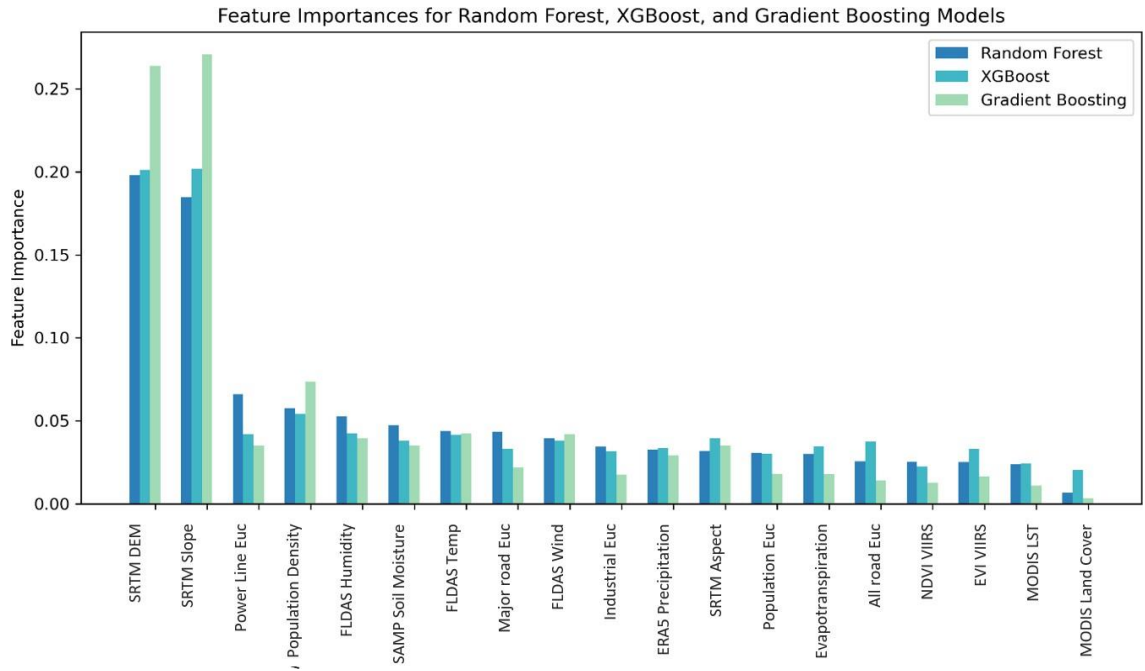

Figure 12: Feature importance analysis for wildfire susceptibility assessment using RF, XGBoost, GBT models based on scenario 2.

Regarding precision, KNN exhibits slightly lower values than RF and XGBoost, with differences of 0.019 and 0.015, respectively. Nonetheless, the KNN algorithm demonstrates the best overall performance in Scenario 2 when considering the results across accuracy metrics. Therefore, KNN is recommended for accurately identifying wildfire-prone areas in this scenario.

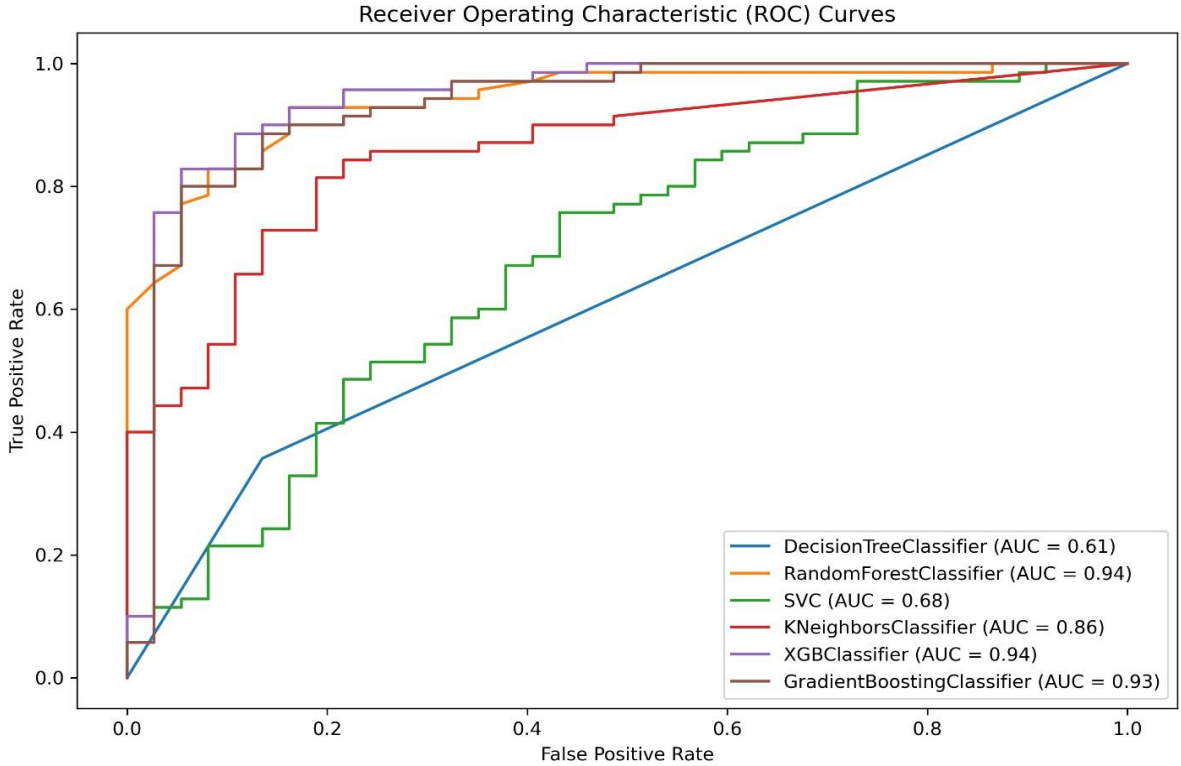

Figure 13: ROC diagram of different algorithms for scenario2.

Table 4: The results of different classification algorithms based on F1-score, recall, precision, and accuracy metrics for scenario 2.

| Metrics/Models | GBT | XGBoost | KNN | SVM | RF | DT |
|---|---|---|---|---|---|---|
| **Accuracy** | 0.542 | 0.598 | 0.682 | 0.402 | 0.495 | 0.533 |
| **Precision** | 0.774 | 0.791 | 0.776 | 0.648 | 0.795 | 0.689 |
| **Recall** | 0.542 | 0.598 | 0.682 | 0.402 | 0.495 | 0.533 |
| **F1-score** | 0.515 | 0.587 | 0.687 | 0.311 | 0.443 | 0.521 |

### 4.2.3 Scenario 3

In Scenario 3, fire and non-fire points, including non-forested areas, were selected without spatial restrictions. Consistent with the previous scenarios, the RF, GBT, and XGBoost models were employed to assess the significance of various parameters in identifying wildfire-prone areas. Figure 14 illustrates the findings.

The analysis reveals that the most significant parameters are elevation, population density, and distance from power lines. Conversely, parameters such as LST and land cover type show minimal importance in identifying these areas, similar to the trends observed in Scenarios 1 and 2.

A key observation from the importance analysis across Scenarios 1, 2, and 3 is that the method of dataset preparation does not influence the ranking of parameter significance. Overall, the results indicate that dataset preparation has a negligible impact on determining the importance of parameters in identifying wildfire-prone areas.

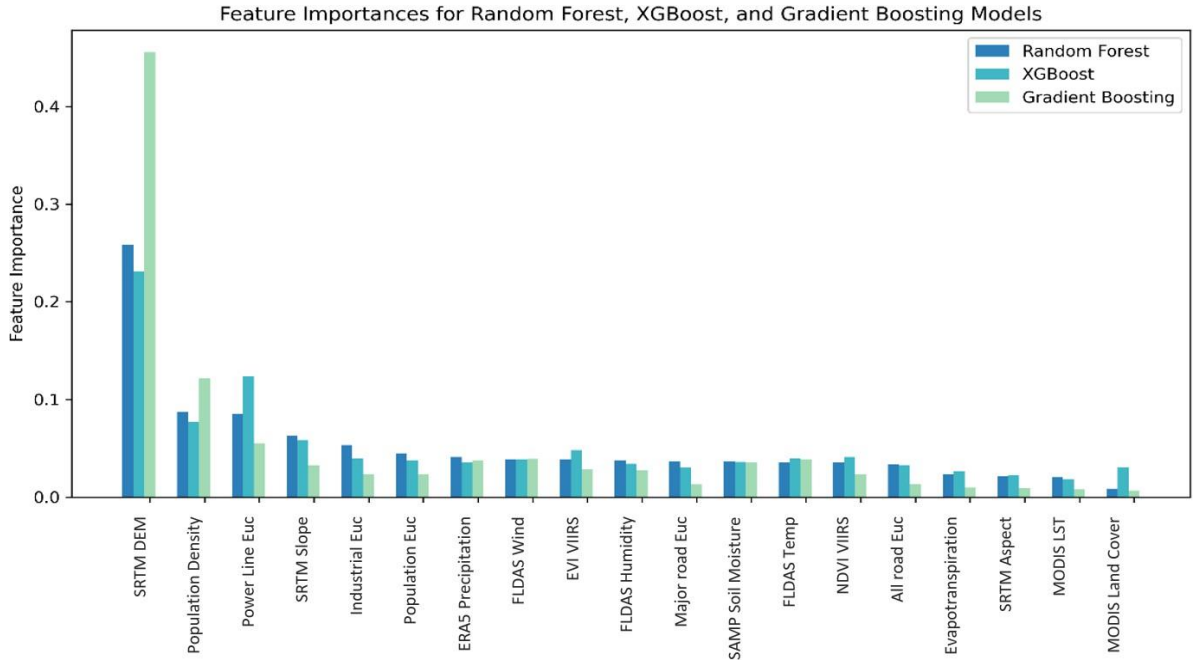

Figure 14: Feature importance analysis for wildfire susceptibility assessment using RF, XGBoost, GBT models based on scenario 3.

Similar to the previous scenarios, the performance of the proposed models was evaluated in this scenario. Figure 15 illustrates the ROC curves for the different models. The results indicate that the RF algorithm achieved the best performance, with an AUC of 97%. It is closely followed by the GBT algorithm, which recorded an AUC of 96%, and the XGBoost algorithm, with an AUC of 95%. In comparison, KNN, SVM, and DT algorithms performed reasonably well, achieving AUCs of 83%, 81%, and 76%, respectively. These findings suggest that RF, GBT, and XGB algorithms are particularly effective for the tasks involved in this scenario, providing a robust approach for identifying wildfire-prone areas.

Table 5 presents a comprehensive overview of various performance metrics for the implemented models in Scenario 3. The GBT algorithm stands out with superior performance across all metrics, achieving an accuracy of 0.900, precision of 0.901, recall of 0.900, and an F1-score of 0.900. Following closely is the RF algorithm, which also delivers competitive results across these metrics.

In contrast, the XGBoost, KNN, SVM, and DT algorithms demonstrate lower performance levels than GBT and RF. Among these, the XGBoost algorithm exhibits the strongest performance, recording an accuracy of 0.863, precision of 0.870, recall of 0.863, and an F1-score of 0.862. Overall, these findings underscore the effectiveness of the GBT and RF algorithms for this classification task, with XGBoost also showing reasonable performance.

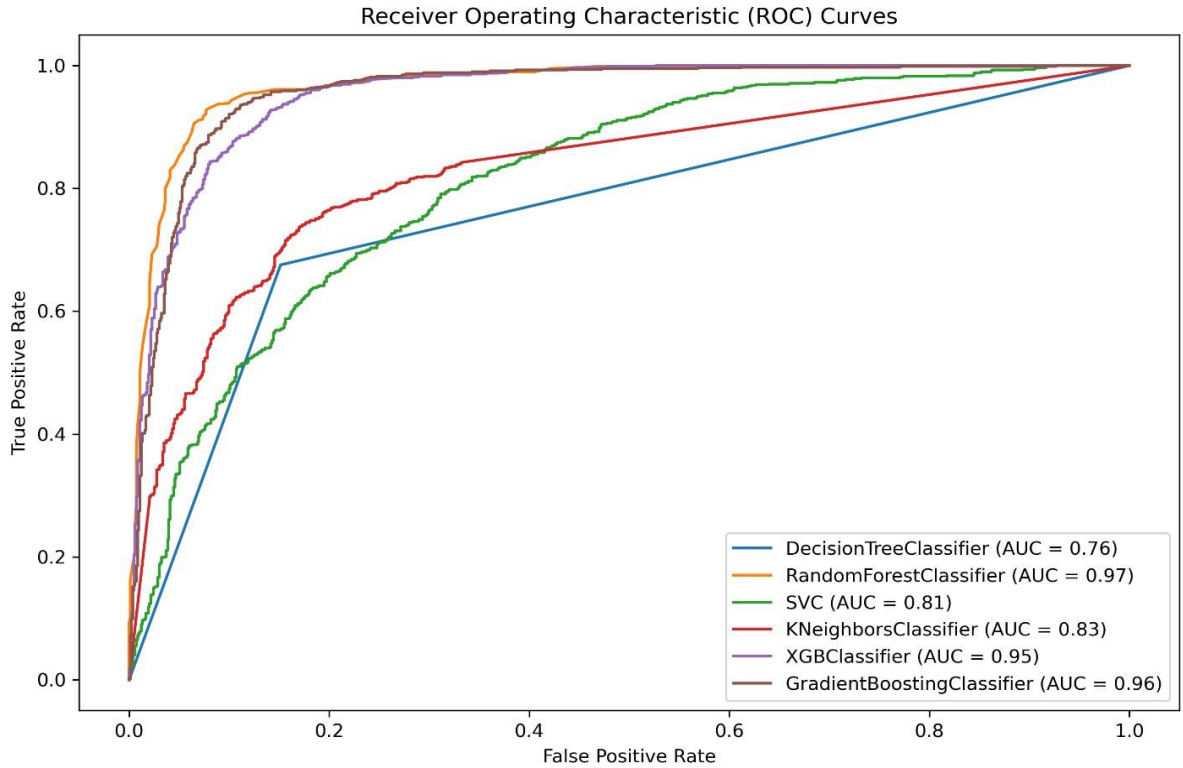

Figure 15: ROC diagram of different algorithms for scenario 3.

Table 5: The performance of different classification algorithms based on F1-score, recall, precision, and accuracy metrics for scenario 3.

| Metrics/Models | GBT | XGBoost | KNN | SVM | RF | DT |
|---|---|---|---|---|---|---|
| **Accuracy** | 0.900 | 0.863 | 0.760 | 0.703 | 0.890 | 0.772 |
| **Precision** | 0.901 | 0.870 | 0.778 | 0.722 | 0.897 | 0.779 |
| **Recall** | 0.900 | 0.863 | 0.760 | 0.703 | 0.890 | 0.772 |
| **F1-score** | 0.900 | 0.862 | 0.754 | 0.695 | 0.889 | 0.770 |

# 5  Discussion

In this study, we evaluated the significance of various input parameters influencing the identification of wildfire-prone areas. Multiple techniques were employed to analyze these parameters, ensuring a comprehensive assessment. The performance of different machine learning models in detecting high risk areas was systematically compared across various sampling scenarios in the study area. This approach aimed to enhance our understanding of which factors most effectively contribute to accurate wildfire prone area identification and to optimize model performance in predicting wildfire susceptibility.

The initial part of this study focuses on analyzing the significance of input parameters and their theoretical connection to the target variable, aiming to validate our proposed method against real-world data. The research also examines the interactions between input parameters and the target variable, assessing how the selection of training and testing data impacts the importance of these parameters.

A critical aspect of this investigation is understanding the role of anthropogenic and climatic factors in identifying wildfire-prone areas. By exploring these influences, the study seeks to provide deeper insights into how these factors affect wildfire susceptibility, leading to a more comprehensive assessment of wildfire risk, particularly in forests.

To investigate this, we first examined the linear correlations among the parameters. The analysis revealed specific relationships within the dataset that can impact the interpretation of model results. For instance, vegetation indices like NDVI and EVI naturally correlate as they both reflect vegetation cover. At the same time, the relationship between humidity and precipitation arises from precipitation being a key source of environmental moisture.

Overall, the correlations are not significant enough to raise concerns about multicollinearity, indicating that the selected features are well-chosen and provide valuable information for identifying wildfire-prone areas. The diversity among these input features allows for a nuanced assessment of human and climatic influences on fire susceptibility, reducing the risk of misleading effects from high correlations.

The NMI and VIF methods were also employed to evaluate the significance of input parameters concerning the target variable. NMI identified soil moisture, humidity, temperature, wind, and precipitation as the most influential parameters. Conversely, anthropogenic factors such as distance from industrial areas, roads, and population centers and indices like land cover type, LST, evapotranspiration, and EVI were deemed less important in this context.

The VIF results generally indicated favorable conditions, showing minimal dependency among input parameters. Furthermore, the analysis of input parameter importance from the perspectives of XGBoost, GBT, and RF algorithms revealed that LST and land cover type had a lesser influence than other parameters. This diminished importance may be attributed to the fact that most samples were collected from similar areas in terms of land type, leading to relatively uniform values for these parameters across the samples. Improving the spatial resolution of the data could enhance their significance in identifying wildfire-prone areas. Higher resolution data may capture more variability, thereby increasing the potential influence of LST and land cover type on wildfire susceptibility.

The varying results from different methods in identifying important parameters indicate that the selected input parameters are well-chosen, as each parameter is deemed significant by at least one method. For example, the NMI method found that human indices were among the less influential parameters, while machine learning models considered these parameters important. The differing results stem from the distinct approaches of these methods. The NMI method measures the degree of mutual dependence between parameters and the target variable based on statistical relationships, revealing how much knowing one parameter informs about another. However, it may not capture the complex, non-linear interactions between parameters and the target variable.

In contrast, machine learning models like XGBoost and RF assess parameter importance by evaluating how each parameter reduces prediction error, effectively capturing these complex relationships. Consequently, parameters that appear less influential in a statistical context (according to NMI) may still be crucial for accurate prediction. For instance, indices such as population density and distance from power transmission lines consistently ranked among the top influential factors in various scenarios. Additionally, the order of parameter importance did not vary based on training and testing sample selection, suggesting that sample collection based on regional characteristics did not significantly impact parameter importance. Moreover, the analysis revealed a notable difference between linear correlations and non-linear relationships identified by XGBoost, GBT, RF, and NMI methods. This indicates that the relationships between most independent variables and the dependent variable are likely non-linear, challenging the effectiveness of linear regression models previously used for modeling wildfire susceptibility.

The parameters influencing the identification of wildfire-prone areas are crucial to consider. Some parameters act as fire-ignition factors, while others contribute to fire spread. Fire-ignition factors include LST, land cover type, soil moisture, temperature, evapotranspiration, Euclidean distances from industrial and populated areas, power transmission lines, primary and secondary roads, and population density. In contrast, fire-propagation factors comprise slope, aspect, EVI, NDVI, precipitation, humidity, and wind speed. Understanding these classifications enhances model accuracy and predictive capabilities. Environmental factors serve as both fire-ignition and fire-propagation factors, whereas human factors, specifically, population density and proximity to powerlines influence ignition. Overall, climatic factors, particularly soil moisture, humidity, and air temperature significantly increase wildfire risk in the study area, while the role of anthropogenic factors should not be overlooked.

The evaluation of various machine learning models for predicting wildfire susceptibility has revealed notable differences in their performance, influenced by the strategies used for selecting training and test samples. In the first scenario, focused on identifying wildfire-prone areas within forested regions, RF, XGBoost, and GBT models showed satisfactory performance compared to traditional methods like DT, KNN, and SVM. A critical factor in this scenario was the selection of non-fire points from varied locations each year, which allowed for improved temporal and spatial generalization.

In the second scenario, designed to eliminate spatial dependence and prevent information leakage, test data were chosen from areas at least 25 km away from training samples. This approach reduced accuracy compared to the first scenario but enhanced generalization to unknown regions. In this scenario, RF, GBT, and XGBoost performed best according to the ROC index, while KNN excelled in other performance metrics. The third scenario aimed to improve model performance in non-forested areas. XGBoost, GBT, and RF performed superiorly, and DT, KNN, and SVM produced satisfactory results. This comprehensive approach facilitated the production of wildfire susceptibility maps for the entire study area. Figure 16 present the wildfire susceptibility maps for 2022 at a 1-kilometer resolution, utilizing the RF, XGBoost, GBT, and KNN methods.

The maps indicate that the southwest region of Iran, specifically the Ilam and Kermanshah provinces within the central Zagros area, is experiencing critical fire crises. This region has been consistently flagged as having a high wildfire risk across multiple algorithms. High-risk conditions are also observed in the Hyrcanian forests of Golestan province in the northeast and the Arasbaran forests in East Azerbaijan province in the northwest.

Key factors contributing to these fires include deliberate arson for charcoal production in Ilam and Kermanshah and significant impacts from consecutive droughts. In northeastern Iran, increased accessibility from main roads and recreational activities that involve lighting fires have heightened wildfire risks. Deliberate arson aimed at converting forests to agricultural land further compounds the issue. Also, the frequent rains in this area lead to a higher incidence of fire initiation through lightning strikes. In the Arasbaran region, the proximity of forests to villages and the accumulation of dry vegetation among trees are critical factors that enhance fire susceptibility. The mountainous terrain with steep slopes also facilitates fire spread. This analysis underscores the pressing need for urgent and effective fire prevention, management, and control measures in these vulnerable regions.

This study not only examines the impact of various parameters on wildfire susceptibility but also introduces seasonal analysis to gain a deeper understanding of how climatic and human factors influence wildfire risk across different times of the year. Seasonal fluctuations significantly influence wildfire risk dynamics, as variables such as temperature, precipitation, and human activities vary markedly between warm and cold seasons. The data was divided into two periods to assess these temporal differences: the cold season (November to April) and the warm season (May to October). The importance of input parameters was evaluated across these seasons to identify temporal patterns and fluctuations in their impact on wildfire risk.

To achieve this, three models—RF, XGBoost, and GBT—were trained using seasonally segmented datasets. The results, summarized in Table 6, reveal that the accuracy of all three models improved during the cold season compared to the scenario using year-round data, whereas a decline in accuracy was observed during the warm season. This variation highlights the effect of seasonal changes on the predictive performance of the models. In the cold season, all three models performed similarly, likely due to more stable climatic conditions providing consistent analysis. Conversely, the RF algorithm outperformed the other models during the warm season.

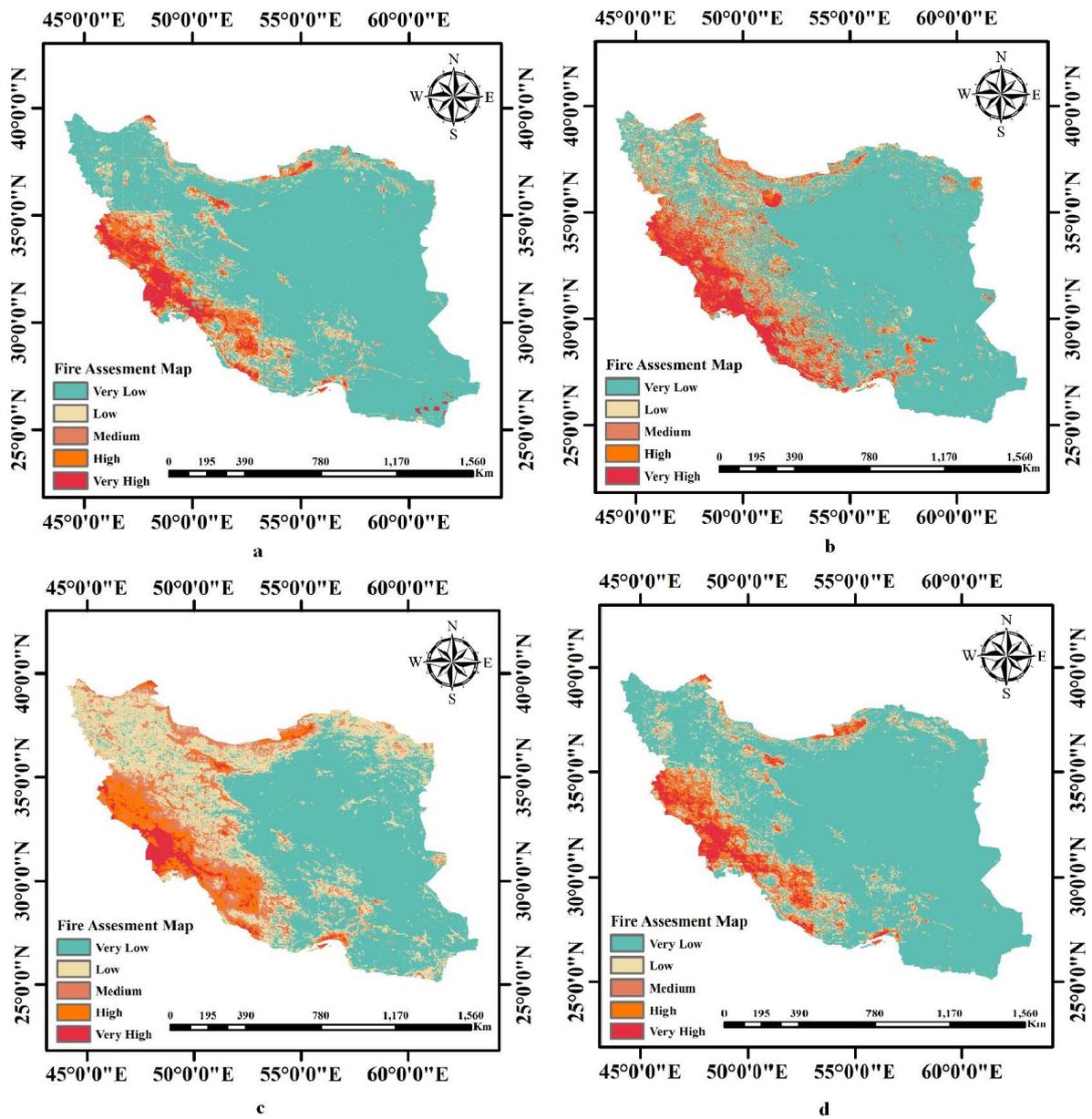

Figure 16:     Wildfire risk maps of Iran prepared by different methods: a) GBT, b) KNN, c) XGBoost, d) RF. Blue areas indicate very low risk and red areas indicate high risk.

Table 6: Impact of seasonal variations on predictive model performance: A comparative analysis of algorithm accuracy in cold and warm seasons.

| Metric/Model | Season | RF | XGBoost | GBT |
|---|---|---|---|---|
| Accuracy | Cold | 0.917 | 0.912 | 0.911 |
| | Warm | 0.801 | 0.738 | 0.738 |
| Precision | Cold | 0.922 | 0.916 | 0.914 |
| | Warm | 0.829 | 0.750 | 0.755 |
| Recall | Cold | 0.917 | 0.912 | 0.911 |
| | Warm | 0.801 | 0.738 | 0.738 |
| F1-score | Cold | 0.917 | 0.912 | 0.911 |
| | Warm | 0.800 | 0.738 | 0.736 |

Figure 17 and Figure 18 depict the importance of parameters during the warm and cold seasons, respectively, based on the aforementioned machine learning algorithms. As shown, human factors play a more significant role than climatic factors in both seasons. However, this distinction becomes less pronounced in the warm season, indicating that the influence of climatic factors is more substantial during the warm season compared to the cold season.

During the cold season, climatic stability (such as solar radiation intensity, snow cover, and frost) accentuates the influence of human factors like proximity to roads and residential areas. In contrast, during the warm season, rising temperatures, reduced precipitation, and changes in vegetation cover make climatic factors more dominant. These findings underscore the importance of incorporating seasonal analysis in wildfire risk management. In the cold season, efforts should focus on controlling human activities, while in the warm season, priority should be given to protecting vegetation and monitoring climatic conditions. Such targeted approaches can improve resource management and enhance preventive strategies.

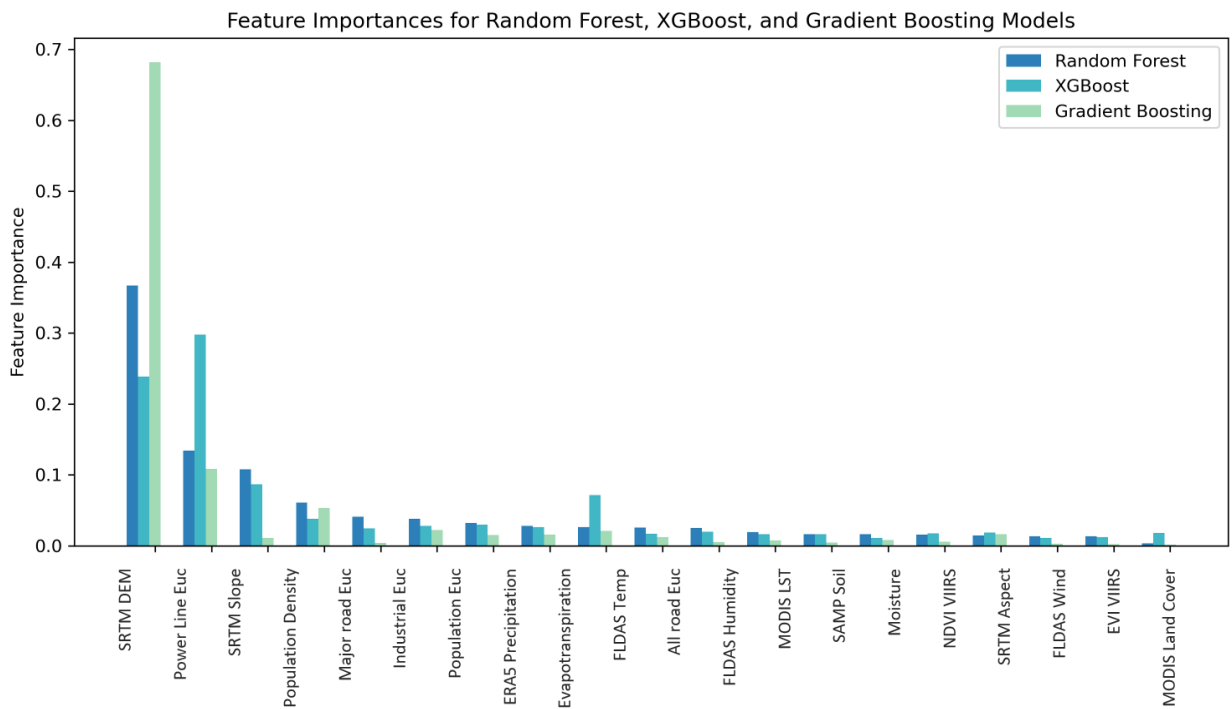

Figure 17: Feature importance analysis for wildfire susceptibility assessment using RF, XGBoost, and GBT models during the cold season.

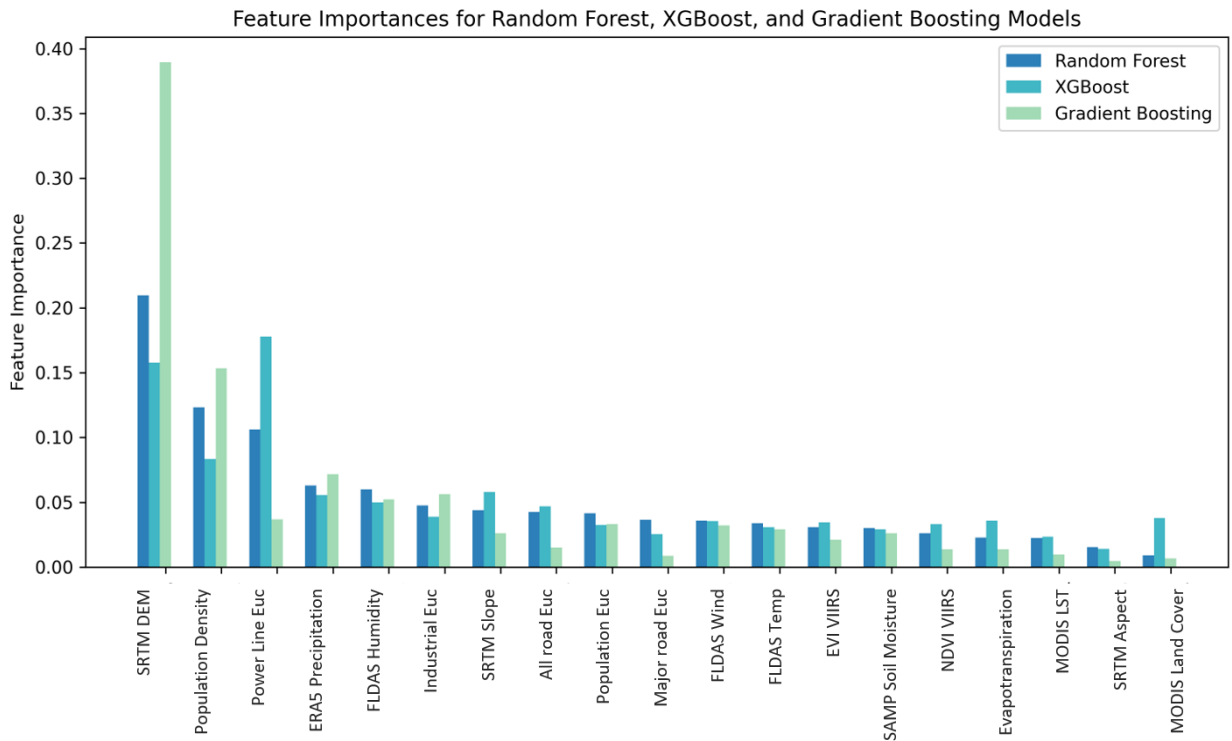

Figure 18: Feature importance analysis for wildfire susceptibility assessment using RF, XGBoost, and GBT models during the warm season.

In this study, we achieved exceptional accuracy in identifying wildfire-prone areas, with our best-performing model reaching an AUC of 0.98. Key advantages of this model include its comprehensive feature set, its application on a national scale across Iran, and its superior predictive power, significantly outperforming previous studies that focused on smaller regions with more limited datasets. As illustrated in Table 7, our model outperformed earlier efforts in identifying wildfire-prone areas.

The first comparison is with Moayedi et al. (2020), who developed a forest fire susceptibility map (FFSM) for Golestan Province, northern Iran, covering cities like Galikesh, Minudasht, and parts of Golestan National Park (Moayedi et al., 2020). Their GA-ANFIS models achieved AUC values of 0.912 for training and 0.850 for testing. Key conditioning factors included elevation, slope, aspect, wind speed, and NDVI. While the models were accurate, they were limited by the small study area and less comprehensive feature set compared to ours.

The second comparison is with Eskandari et al. (2020) in Golestan Province, who evaluated both individual and hybrid machine learning algorithms (Eskandari et al., 2020). Their best-performing hybrid model achieved an AUC of 0.96 but, like other studies, was restricted to a smaller area and a narrower set of input features. Their model used fire occurrence data from NASA's FIRMS, a DEM from ASTER-GDEM, and features such as slope angle, plan curvature, and topographic wetness index. In contrast, our study incorporated a more comprehensive range of parameters, improving predictive accuracy, particularly in high-risk areas, and enabling fire susceptibility assessment across a more extensive and diverse landscape.

Table 7: Comparing the results of this investigations with similar studies of wildfire susceptibility assessment in Iran

| Study | Model | AUC (Test) | Features | Study Area |
|-------|-------|------------|----------|------------|
| **Our Study** | GBT, Random Forest | **0.98** | Topographic, vegetation, human, environmental | Iran (national scale) |
| Moayedi et al. (2020) [70] | GA-ANFIS | 0.850 | Environmental, topographic, climate, FR conditioning factors | Golestan Province (Galikesh and Minudasht) |
| Eskandari et al. (2020) [71] | Hybrid Model | 0.96 | Environmental, climate, vegetation, fire occurrence data, DEM, distance measures | Golestan Province, Iran |

# 1 Conclusion

This research adopted a comprehensive and multifaceted analytical approach to assess the impact of anthropogenic and climatic factors on wildfire risk in Iran. For this purpose, cloud-based platforms such as GEE and Overpass Turbo were employed. These platforms allowed immediate access to large-scale remote sensing data and spatial analysis tools without the need for advanced user-side infrastructure, as all processing and data storage were managed on cloud servers. The continuous updates and accessibility of these platforms' temporal and spatial analytical tools made the process more efficient, enabling the more accurate and faster identification of high-risk areas, which is critical for implementing proactive wildfire management measures. The methodology included the use of traditional statistical techniques as well as advanced machine learning models. Initial evaluations revealed that human and climatic factors significantly influence wildfire risk, though their relative importance depends on data collection methods and analysis techniques. Notably, the relationships between most parameters and wildfire risk were found to be nonlinear, rendering linear regression models ineffective for accurate prediction. To address this limitation, advanced machine learning models were utilized, which were tested under various data collection scenarios to provide reliable results across different spatial and temporal contexts. These models produced accurate maps of high-risk areas such as the central Zagros region, the eastern Hyrcanian forests, and Arasbaran. One of this research's key findings was identifying population density as the most critical factor in wildfire occurrence, emphasizing the combined effects of human and environmental factors on wildfire risk. Moreover, this study also investigated the influence of climatic and anthropogenic factors on wildfire risk across various seasons. The impact of these parameters was assessed separately for warm and cold seasons.

The findings underscored the prominent role of human-related factors, particularly population density and proximity to powerlines, which exhibited a stronger correlation with wildfire risk during the seasonal analysis compared to climatic variables. This seasonal evaluation provided a more comprehensive understanding of the dynamic nature of wildfire risk and emphasized the critical importance of incorporating seasonal fluctuations into future risk assessments. Additionally, this study created a 1-kilometer resolution wildfire risk map for the entire country, which can serve as a valuable tool for predicting wildfire risk in the coming year. Future research should employ more advanced machine learning algorithms, including deep neural networks, to better model the complex interactions between human and environmental factors. These methods can enhance prediction accuracy and provide more reliable insights for wildfire risk management.

## Acknowledgments

We sincerely thank all individuals and organizations contributing to this research. Thanks to the Statistical Center of Iran for their indispensable data and insights. We also acknowledge the significant contributions of various data providers, including NASA for providing data from various sensors, including SRTM, VIIRS, MODIS, and SMAP, and the outputs of FLDAS, ECMWF for ERA5 data, and the OpenStreetMap Foundation for providing various geospatial layers.

## Data availability

All Codes, datasets and results associated with this research are publicly available at our GitLab repository https://gitlab.com/isfahan-lab/forest-fire-prediction.